\documentclass[11pt]{article}
\usepackage[preprint]{acl}
\usepackage{times}
\usepackage{latexsym}
\usepackage[T1]{fontenc}
\usepackage[utf8]{inputenc}
\usepackage{microtype}
\usepackage{fvextra}
\usepackage{inconsolata}
\usepackage{amsmath}
\usepackage{graphicx}
\usepackage{booktabs}
\usepackage{threeparttable}
\usepackage{placeins}
\usepackage{float}
\DefineVerbatimEnvironment{PromptBox}{Verbatim}{breaklines=true,breakanywhere=true,fontsize=\footnotesize}

\title{A, B, C, or Stable? Accuracy and Order Sensitivity Diverge in LLM Multiple-Choice Evaluation}

\author{
  Karl Hanna \\
  Queen's University Belfast \\
  \texttt{khanna28@qub.ac.uk}
  \And
  Chen Feng \\
  Queen's University Belfast \\
  \texttt{c.feng@qub.ac.uk}
}

\begin{document}
\maketitle
\begin{abstract}

Multiple-choice question (MCQ) benchmarks are widely used to evaluate large language models, yet a single reported accuracy number is not enough to distinguish a model's latent ability from order sensitivity. Debiasing methods tend to be judged on one axis, which implicitly assumes that fixing order sensitivity and improving accuracy go together. We test this assumption with two proprietary models and two open-weight models on MMLU and ARC-Challenge, across four families of strategies. We measure order sensitivity with flip rate, capturing per-question instability, and recall standard deviation (RStd), capturing aggregate positional imbalance. We find accuracy and order sensitivity do not move together consistently: two-stage prompting reduces both flip rate and RStd for GPT on MMLU without a significant change in accuracy. The two sensitivity metrics also diverge from each other. Adding reasoning to direct answering leaves Llama's MMLU flip rate statistically indistinguishable while RStd falls sharply. A $2\times2$ ablation across generation-then-matching, crossing option visibility with matcher type, shows that withholding options mainly harms semantic matching, which achieves zero flip rate by construction, yet loses significantly in accuracy; the effect is smaller when using an LLM matcher. We recommend reporting flip rate and RStd along with accuracy when evaluating MCQ strategies.

\end{abstract}
\section{Introduction}

Multiple-choice question (MCQ) benchmarks are one of the wide-spread evaluation formats \cite{hendrycks2021mmlu, clark2018arc}, since they are inexpensive to run compared to other formats and can be scored automatically. However, the score is only as trustworthy as the evaluation is robust. A model's accuracy on an MCQ benchmark does not distinguish knowledge from sensitivity to option order \cite{pezeshkpour-hruschka-2024-large, zheng2023large, wang-etal-2025-llms-may}. 

Existing work has proposed numerous methods for reducing option-order effects and improving MCQ performance, but less is known about how changes in order sensitivity relate to changes in accuracy. A strategy may make a model's predictions more uniform across option permutations, without improving accuracy, while another may improve accuracy without reducing order-sensitivity. Understanding this relationship is important because accuracy and robustness metrics capture different properties of MCQ evaluation, and improvement in one may not imply improvement in the other.

In this paper, we test a range of MCQ answering strategies across four models, comprising two proprietary models (GPT-4.1 mini and Claude Haiku 4.5) and two open-weight models served over API (Llama-3.1-8B-Instruct at FP8 and Qwen 2.5 7B). We evaluate each model using a range of methods targeting distinct components of the MCQ answering process. We test direct MCQ, cyclic permutation, two-stage strategies, and reasoning variants, alongside decomposing generation-then-matching strategies into a $2\times2$ grid to isolate which factor drives their failure, as well as an approach in which the model never sees the options as distinct choices. We provide a detailed breakdown of Flip rate, RStd, and accuracy. We additionally run a subset of questions on locally served models to assess consistency with the hosted-model results.

Together, these analyses separate question instability (flip rate), positional imbalance (RStd), and raw performance (accuracy). We find that accuracy and order sensitivity can move independently: on MMLU, two-stage prompting substantially reduces both metrics for GPT without a significant change in accuracy. A $2\times2$ decomposition of generation-then-matching shows that its accuracy losses come mainly from hiding the options when matching is semantic. The two sensitivity metrics can also disagree, with reasoning leaving Llama's flip rate unchanged while sharply reducing RStd. Our results therefore motivate evaluating MCQ strategies jointly across accuracy and order sensitivity rather than treating any single isolated metric as sufficient.
\section{Related Work}

\paragraph{Multiple-choice brittleness and option-order sensitivity.}
Several papers show that large language models are not robust multiple-choice reasoners, even when their benchmark accuracy looks strong. \citet{zheng2023large} find that LLMs behave as unreliable multiple-choice selectors, which motivates methods that reduce sensitivity in the final selection step. \citet{zheng2023large} also introduce RStd, the standard deviation of per-option recall, which we use. \citet{pezeshkpour-hruschka-2024-large} show that performance changes substantially when the options are reordered, which suggests that multiple-choice evaluation may be measuring sensitivity to formatting rather than stable underlying knowledge.

\paragraph{Debiasing and robustness-oriented prompting methods.}
Several approaches try to make multiple-choice evaluation more robust by modifying the prompting or selection procedure. PriDe~\citep{zheng2023large} estimates a prior from a small calibration set derived from the test set, and logarithmically subtracts it from the prediction distribution. They also introduce cyclic permutation combating the high cost of using full permutations, though we use it aggregated over a majority vote since we have no full logit access. Chain-of-thought prompting \citep{kojima2022large, wei2022chain} is a natural strategy to try, which we also test variants of independently, but \citet{zheng2023large} find that it does not meaningfully reduce selection bias.

\paragraph{Open-style answering and answer matching.}
Another line of work questions whether forced-choice evaluation is the right interface for measuring model knowledge at all. \citet{myrzakhan2024open} argue for moving from purely multiple-choice formats toward open-style questions in leaderboard-style evaluation, since multiple-choice constraints can distort what is actually being measured. \citet{chandak2025answer} go further and show that answer matching outperforms standard multiple-choice evaluation, which suggests that free-form responses give a more faithful picture of model competence than letter selection alone, and outperforms Direct Answering in terms of human alignment.

\paragraph{Isolated per-option scoring.}
A separate line of work removes option-order effects structurally rather than correcting them after they occur. Set-Based Prompting~\citep{mcilroy2024order} modifies the attention mask and positional encoding so that option order cannot affect the output. Its effect on accuracy is small, typically remaining within the variation caused by reordering under ordinary prompting: order invariance is achieved, but accuracy is essentially unchanged. \citet{balepur2024artifacts} evaluate each option separately under both question-present and choices-only settings, with their question-present setup closely resembling independent hypothesis scoring but using a binary correctness judgement rather than a numeric score. \citet{chandak2025answer} test a related multiple-choice verification setup, where each choice is shown separately and independently judged as correct or incorrect. Together, these findings suggest a potential cost to option isolation: it prevents the model from directly comparing the available choices.

\paragraph{Positioning of this work.}
This paper sits at the intersection of these strands. Prior work explores multiple debiasing methods, alternative answering strategies, and whether MCQ evaluation is even suitable in the first place. \citet{zheng2023large} find that removing option IDs reduced selection bias but usually lowered accuracy, and \citet{mcilroy2024order} achieved order invariance with essentially unchanged accuracy. For those papers, these were secondary findings, each about a single intervention, reported in passing rather than examined directly. In this paper, we instead study how these different aspects of MCQ evaluation relate to each other, particularly accuracy and flip rate and RStd (order sensitivity).
\section{Methodology}

\subsection{Strategies}
We compare a range of MCQ strategies that either modify individual components of the evaluation process or take qualitatively different approaches to mitigating positional sensitivity. We begin with direct MCQ, where the model is presented with the question and labelled options, together with a reasoning-enabled \citep{kojima2022large, wei2022chain} variant that explicitly instructs the model to reason before answering. We then evaluate two-stage prompting, where the model first produces a preliminary free-text answer without access to the options, and then an LLM matches this output to one of the presented choices, again with a reasoning-enabled counterpart. To isolate the components responsible for its behaviour, we use a \(2\times2\) diagnostic crossing option visibility during generation to produce a free-text response with LLM-based versus semantic matching. The names of all four methods in this diagnostic are Hidden+LLM (referred to as two-stage prompting outside the diagnostic), Visible+LLM, Hidden+Semantic and Visible+Semantic. The semantic matching we use here involves exact matching, then falls back to containment, then embedding similarity with additional details in Appendix ~\ref{appendix:semantic}.

We additionally evaluate cyclic permutation \citep{zheng2023large}, where all cyclic rotations of the answer options are queried and aggregated by majority vote, together with a reasoning-enabled variant (we aggregate over majority votes rather than probability over the full logits). Independent Hypothesis scoring (IHS) instead evaluates each option independently by presenting the question with one option at a time, extracting a score for each from 0--100, and selecting the highest-scoring option, with ties being broken deterministically over option content. Cyclic permutation provides an established positional-bias mitigation approach against which the other strategies can be contextualized.

Full prompt templates and implementation details are given in Appendix~\ref{appendix:prompts}.

\subsection{Experimental Setup}

We evaluate GPT-4.1 mini, Claude Haiku 4.5, Llama 3.1-8B Instruct, and Qwen 2.5 7B Instruct through hosted APIs. All generation has a cap of 1024 tokens and a temperature of 0.

Both benchmarks are English-language multiple-choice datasets. MMLU spans 57 subjects across the humanities, social sciences, and STEM, while ARC-Challenge comprises grade-school science questions \citep{hendrycks2021mmlu,clark2018arc}. We use 1,140 sampled MMLU test questions and all 1,172 ARC-Challenge test questions for evaluation on all the different model $\times$ method runs. We used the ARC-Challenge test set specifically, and used 20 questions from each of the 57 subjects of MMLU, randomly sampled with seed 42. A subset is re-run on locally served models available in the Appendix~\ref{appendix:results}. 

\subsection{Evaluation}
We use two different accuracy metrics: end-to-end accuracy (E2E) where the denominator is the full question set, regardless of any failures such as generation-length limit, and conditional accuracy, where the denominator is only the scorable outputs. We use these to see how accurately a method performs when producing a scorable output, but also use E2E since a much lower E2E accuracy is still a property of the method even if it has a high conditional accuracy. Next, we use flip rate to measure how volatile a model's answer is when the option order is permuted. An answer is considered flipped when the choice, mapped back to the canonical option content, is not identical across all cyclic rotations. By construction, Hidden+Semantic is invariant to option positions. IHS has no rotations, so it has no measured flip rate. Lastly, we use Recall Standard Deviation (RStd) \citep{zheng2023large}: the standard deviation of the model's recall across gold-answer positions. Unlike the original single-ordering version, we pool observations across all cyclic rotations, so that under complete parsing, every question contributes once to every gold position, and each position's recall is computed over the same set of questions. This accounts for differences in question difficulty across positions, so any remaining imbalance reflects how the model responds to position itself. IHS is the exception: since it has no rotations, its RStd is computed from the single canonical ordering and can reflect question composition rather than positional bias.

We use several statistical methods to quantify uncertainty and compare methods. For accuracy, we report 95\% Clopper--Pearson confidence intervals. Since RStd, flip rate, and deltas are derived statistics rather than a single boolean outcome, we estimate the uncertainty using 10,000 question-level bootstrap resamples and report 95\% percentile intervals. Finally, we use McNemar's test with Holm correction to compare paired correctness outcomes between each method and the corresponding direct-MCQ baseline.
\section{Results}
\label{sec:results}

\subsection{Accuracy}

Direct MCQ with reasoning achieves higher point-estimate accuracy in 7/8 cells, with the sole exception of Claude on ARC, where it is slightly worse, although some of the 95\% CIs for the paired accuracy differences include zero. Overall, cyclic permutation with reasoning instructed is the strongest method by point-estimate accuracy, scoring the highest point estimate in all cells except GPT$\times$MMLU, and improving over its non-reasoning counterpart, cyclic permutation, in all eight cells.

However, we can see that reasoning does not consistently improve every strategy. Looking at reasoning two-stage, we can see that it is much more mixed. On ARC, its point-estimate accuracy is lower for GPT, Claude and Qwen, and higher for Llama. More interestingly, looking at IHS, we can see that it does not provide a consistent improvement even though it is option-order invariant by construction. Full accuracy results are reported in Tables~\ref{tab:accuracy-mmlu} and~\ref{tab:accuracy-arc} in the Appendix.

Conditional and end-to-end accuracy are nearly identical for most conditions, indicating that unscored outputs are generally limited. The main exceptions are Hidden+Semantic and Qwen Visible+Semantic, where substantial numbers of unscored questions create large gaps between conditional and end-to-end accuracy. Full conditional-accuracy results are reported in Tables~\ref{tab:conditional-mmlu} and~\ref{tab:conditional-arc} in the Appendix.

In addition, IHS exhibits substantially different behaviour across models. For the four models on MMLU, 4.9\% of Claude, 12.7\% of GPT, 18.8\% of Qwen, and 21.2\% of Llama questions have correctness determined by tie-breaking, dropping to 2.8\%, 5.7\%, 11.2\%, and 15.4\% on ARC, respectively. Approximately one fifth of Qwen and Llama MMLU questions therefore have correctness determined purely by tie-breaking. GPT, Qwen and Llama overwhelmingly output scores in multiples of 5, with Claude using much finer scores and correspondingly having lower tie rates. Other notable observations are that Qwen and Llama errors are only about one third lost ties, with about two thirds genuinely misranked; IHS cannot faithfully evaluate relational options such as ``all of the above'' in isolation, and 78 MMLU questions contain this type of option, with IHS trailing Direct by 12--23 pp on them across all four models. Most notably, among the 41 questions whose correct answer is a relational option, 37 place that gold answer at position D. GPT's D-position accuracy rises from 77.4 to 82.6 once those questions are excluded. This provides concrete evidence that nonzero single-ordering RStd does not necessarily imply positional bias. A full breakdown of IHS tie rates, score behaviour, and relational-option performance is reported in Appendix Table~\ref{tab:ihs-score-analysis}.

\subsection{Order Sensitivity}
\label{subsec:order_sensitivity}

Firstly, one of the most stand-out results is that RStd and flip rate are not interchangeable. Across all cells, we can see that both metrics do not consistently move together, with some cells having high flip rates and RStd, others having low flip rates and RStd, while others are more irregular. Together, they provide complementary views of order sensitivity.

For example, Llama scores a flip rate of 54.74 and an RStd of 5.56 on Direct MCQ, while scoring a statistically indistinguishable flip rate of 53.25 on Reasoning-Direct, with RStd dropping sharply to 0.81. This shows that a model can remain highly permutation-sensitive while its position-specific recall becomes much more uniform.

Two-stage approaches sometimes reduce flip rate substantially, especially on MMLU, but this does not generalize cleanly to ARC. In addition, reasoning-enabled counterparts are not uniformly stabilizing either; in some comparisons, both Flip and RStd are statistically indistinguishable from their non-reasoning counterparts. In terms of flip rate, Llama is the most sensitive model, with Qwen coming in second and GPT and Claude lower. ARC also generally has lower flip rates and RStd values compared to MMLU. Full order-sensitivity results are reported in Tables~\ref{tab:order-mmlu} and~\ref{tab:order-arc} in the Appendix.

\subsection{Accuracy and Order Sensitivity Relative to Direct}
\label{subsec:accuracy_sensitivity}

Tables~\ref{tab:direct-delta-mmlu} and~\ref{tab:direct-delta-arc} summarize changes relative to Direct. Accuracy, flip rate, and RStd do not move together consistently, with several different patterns appearing across metrics, models, and methods. The corresponding paired 95\% CIs are reported in Tables~\ref{tab:delta-unc-mmlu} and~\ref{tab:delta-unc-arc} in the Appendix.

\begin{table*}[t]
\centering
\footnotesize
\setlength{\tabcolsep}{3pt}
\caption{Accuracy and option-order sensitivity relative to Direct MCQ on MMLU (API models). The Direct row reports absolute Accuracy (\%) / Flip (\%) / RStd. All other rows report $\Delta$Accuracy / $\Delta$Flip / $\Delta$RStd relative to the same model's Direct row; accuracy and flip differences are in percentage points. Negative $\Delta$Flip or $\Delta$RStd means lower order sensitivity or positional imbalance than Direct. Flip: share of questions whose parsed choice (mapped to the original option order) is not identical across all cyclic option rotations. RStd: standard deviation of position-specific recall, over 4-option questions. Accuracy is end-to-end (unscored questions count as incorrect); $\Delta$Accuracy is over all questions. $\Delta$Flip and $\Delta$RStd are computed on the questions for which the metric is defined under both the condition and Direct. Differences are computed from unrounded values.}
\label{tab:direct-delta-mmlu}
\begin{tabular}{@{}lcccc@{}}
\toprule
Method & GPT-4.1 mini & Claude Haiku 4.5 & Qwen2.5-7B & Llama-3.1-8B \\
\midrule
Direct & 81.84\,/\,20.26\,/\,2.02 & 84.47\,/\,22.54\,/\,0.42 & 74.30\,/\,31.23\,/\,2.29 & 59.12\,/\,54.74\,/\,5.56 \\
\midrule
Reasoning Direct & +7.28\,/\,$-$7.19\,/\,$-$1.41 & +2.28\,/\,$-$1.07\,/\,+0.27 & +0.96\,/\,+2.54\,/\,+1.30 & +6.49\,/\,$-$1.49\,/\,$-$4.75 \\
Two-stage & $-$0.26\,/\,$-$7.63\,/\,$-$1.57 & $-$1.40\,/\,$-$5.77\,/\,+1.03 & $-$5.79\,/\,+0.53\,/\,+0.93 & +0.26\,/\,$-$5.27\,/\,+3.27 \\
Reasoning Two-stage & +3.95\,/\,$-$9.30\,/\,$-$1.24 & +2.28\,/\,$-$7.56\,/\,+1.44 & $-$3.95\,/\,+0.53\,/\,+1.06 & +0.53\,/\,$-$7.73\,/\,+3.82 \\
Visible+LLM & +2.37\,/\,$-$3.51\,/\,$-$0.73 & $-$1.49\,/\,$-$2.06\,/\,+0.48 & $-$1.05\,/\,+0.61\,/\,+0.22 & +3.95\,/\,$-$0.96\,/\,$-$1.31 \\
Visible+Semantic & +1.58\,/\,$-$3.33\,/\,$-$0.74 & +1.75\,/\,$-$6.50\,/\,+0.39 & $-$15.09\,/\,$-$2.06\,/\,$-$0.07 & $-$0.09\,/\,$-$3.08\,/\,$-$1.48 \\
\bottomrule
\end{tabular}
\par\vspace{2pt}
{\scriptsize 95\% paired question-bootstrap CIs for every difference are in Table~\ref{tab:delta-unc-mmlu}.}
\end{table*}

\begin{table*}[t]
\centering
\footnotesize
\setlength{\tabcolsep}{3pt}
\caption{Accuracy and option-order sensitivity relative to Direct MCQ on ARC-Challenge (API models). The Direct row reports absolute Accuracy (\%) / Flip (\%) / RStd. All other rows report $\Delta$Accuracy / $\Delta$Flip / $\Delta$RStd relative to the same model's Direct row; accuracy and flip differences are in percentage points. Negative $\Delta$Flip or $\Delta$RStd means lower order sensitivity or positional imbalance than Direct. Flip: share of questions whose parsed choice (mapped to the original option order) is not identical across all cyclic option rotations. RStd: standard deviation of position-specific recall, over 4-option questions. Accuracy is end-to-end (unscored questions count as incorrect); $\Delta$Accuracy is over all questions. $\Delta$Flip and $\Delta$RStd are computed on the questions for which the metric is defined under both the condition and Direct. Differences are computed from unrounded values.}
\label{tab:direct-delta-arc}
\begin{tabular}{@{}lcccc@{}}
\toprule
Method & GPT-4.1 mini & Claude Haiku 4.5 & Qwen2.5-7B & Llama-3.1-8B \\
\midrule
Direct & 95.82\,/\,6.06\,/\,0.18 & 94.54\,/\,8.36\,/\,0.67 & 89.42\,/\,14.16\,/\,0.89 & 77.22\,/\,29.95\,/\,2.26 \\
\midrule
Reasoning Direct & +0.34\,/\,$-$2.30\,/\,$-$0.06 & $-$0.26\,/\,+2.65\,/\,$-$0.05 & +1.54\,/\,+0.68\,/\,+1.23 & +6.40\,/\,+1.28\,/\,$-$0.89 \\
Two-stage & $-$1.19\,/\,$-$0.17\,/\,+0.21 & $-$2.05\,/\,+0.94\,/\,+0.07 & $-$5.38\,/\,+5.80\,/\,+1.75 & $-$7.25\,/\,+6.31\,/\,+5.29 \\
Reasoning Two-stage & $-$1.54\,/\,$-$0.68\,/\,+0.29 & $-$2.39\,/\,+1.11\,/\,+0.27 & $-$8.70\,/\,+12.46\,/\,+4.50 & $-$3.58\,/\,+6.23\,/\,+6.59 \\
Visible+LLM & +0.26\,/\,$-$0.17\,/\,+0.25 & $-$0.26\,/\,$-$2.57\,/\,$-$0.28 & $-$0.17\,/\,+0.43\,/\,+0.41 & +0.68\,/\,+2.82\,/\,+1.09 \\
Visible+Semantic & +0.09\,/\,+0.09\,/\,+0.28 & +0.00\,/\,$-$2.82\,/\,$-$0.30 & $-$10.58\,/\,$-$0.61\,/\,+0.65 & $-$1.28\,/\,+4.52\,/\,+0.82 \\
\bottomrule
\end{tabular}
\par\vspace{2pt}
{\scriptsize 95\% paired question-bootstrap CIs for every difference are in Table~\ref{tab:delta-unc-arc}.}
\end{table*}

Zooming in on GPT$\times$MMLU$\times$Two-stage, the 95\% CI for the change in accuracy includes zero, while both flip rate and RStd substantially decrease. It is evident that order sensitivity can improve without a corresponding accuracy gain. 

Looking at the cells overall, we see other irregular patterns. Qwen$\times$MMLU$\times$Visible+Semantic drops sharply in accuracy while remaining statistically indistinguishable from Direct on both order-sensitivity metrics. Qwen$\times$ARC$\times$Reasoning Direct improves in accuracy while RStd worsens, with no distinguishable change in flip rate. All these varying examples demonstrate that accuracy, per-question permutation sensitivity, and positional imbalance capture distinct properties in MCQ answering. Improving one does not necessarily imply an improvement in the others, and vice versa.

Paired correctness comparisons using exact McNemar tests with Holm correction are reported in Tables~\ref{tab:mcnemar-mmlu} and~\ref{tab:mcnemar-arc} in the Appendix.

\subsection{The \texorpdfstring{$2\times2$}{2x2} Diagnostic}
\label{subsec:grid}

Moving onto the $2\times2$ diagnostic, which varies option visibility along with matcher type, there is a clean observation. Visible+LLM, Visible+Semantic, and Hidden+LLM are generally within the same ballpark of accuracy, while Hidden+Semantic shows sharp losses.

\begin{table*}[t]
\centering
\footnotesize
\setlength{\tabcolsep}{4pt}
\caption{Two-stage $2\times2$ diagnostic on MMLU, API models. Rows: option visibility in the answer-generating prompt (hidden/visible) $\times$ matcher (LLM/semantic). Each cell, top to bottom: Accuracy (\%) with 95\% Clopper--Pearson CI; Flip (pp) and RStd (pp) with 95\% question-bootstrap CIs; scored/total where fewer than all questions were scored. Hidden+LLM is the Two-stage condition; Visible+Semantic is the Text-extraction (cascade) condition.}
\label{tab:2x2-mmlu}
\begin{tabular}{@{}lcccc@{}}
\toprule
Condition & GPT-4.1 mini & Claude Haiku 4.5 & Qwen2.5-7B & Llama-3.1-8B \\
\midrule
Hidden+LLM & \begin{tabular}[t]{@{}c@{}}81.58 [79.20, 83.79] \\ 12.63 [10.70, 14.56] \\ 0.45 [0.20, 1.06] \\ {\scriptsize 1139/1140 scored}\end{tabular} & \begin{tabular}[t]{@{}c@{}}83.07 [80.76, 85.20] \\ 16.51 [14.39, 18.61] \\ 1.45 [0.91, 2.13] \\ {\scriptsize 1128/1140 scored}\end{tabular} & \begin{tabular}[t]{@{}c@{}}68.51 [65.72, 71.20] \\ 31.75 [29.12, 34.47] \\ 3.22 [2.42, 4.11] \\ {\scriptsize 1139/1140 scored}\end{tabular} & \begin{tabular}[t]{@{}c@{}}59.39 [56.47, 62.25] \\ 49.38 [46.44, 52.28] \\ 8.81 [7.76, 9.91] \\ {\scriptsize 1136/1140 scored}\end{tabular} \\
Hidden+Semantic & \begin{tabular}[t]{@{}c@{}}46.67 [43.74, 49.61] \\ 0.00 [0.00, 0.00] \\ 0.00 [0.00, 0.00] \\ {\scriptsize 985/1140 scored}\end{tabular} & \begin{tabular}[t]{@{}c@{}}44.74 [41.82, 47.68] \\ 0.00 [0.00, 0.00] \\ 0.00 [0.00, 0.00] \\ {\scriptsize 910/1140 scored}\end{tabular} & \begin{tabular}[t]{@{}c@{}}40.70 [37.83, 43.62] \\ 0.00 [0.00, 0.00] \\ 0.00 [0.00, 0.00] \\ {\scriptsize 973/1140 scored}\end{tabular} & \begin{tabular}[t]{@{}c@{}}37.37 [34.55, 40.25] \\ 0.00 [0.00, 0.00] \\ 0.00 [0.00, 0.00] \\ {\scriptsize 941/1140 scored}\end{tabular} \\
Visible+LLM & \begin{tabular}[t]{@{}c@{}}84.21 [81.96, 86.28] \\ 16.75 [14.65, 18.95] \\ 1.29 [0.79, 1.94]\end{tabular} & \begin{tabular}[t]{@{}c@{}}82.98 [80.67, 85.12] \\ 20.59 [18.25, 22.95] \\ 0.92 [0.40, 1.72] \\ {\scriptsize 1111/1140 scored}\end{tabular} & \begin{tabular}[t]{@{}c@{}}73.25 [70.57, 75.80] \\ 31.84 [29.12, 34.56] \\ 2.51 [1.81, 3.35]\end{tabular} & \begin{tabular}[t]{@{}c@{}}63.07 [60.19, 65.88] \\ 53.77 [50.88, 56.67] \\ 4.25 [3.28, 5.35] \\ {\scriptsize 1139/1140 scored}\end{tabular} \\
Visible+Semantic & \begin{tabular}[t]{@{}c@{}}83.42 [81.13, 85.54] \\ 16.93 [14.82, 19.21] \\ 1.28 [0.78, 1.93] \\ {\scriptsize 1138/1140 scored}\end{tabular} & \begin{tabular}[t]{@{}c@{}}86.23 [84.09, 88.18] \\ 16.04 [13.92, 18.15] \\ 0.85 [0.43, 1.50] \\ {\scriptsize 1133/1140 scored}\end{tabular} & \begin{tabular}[t]{@{}c@{}}59.21 [56.29, 62.08] \\ 29.36 [26.60, 32.12] \\ 2.14 [1.35, 3.15] \\ {\scriptsize 921/1140 scored}\end{tabular} & \begin{tabular}[t]{@{}c@{}}59.04 [56.12, 61.91] \\ 51.58 [48.59, 54.53] \\ 4.10 [3.14, 5.21] \\ {\scriptsize 1086/1140 scored}\end{tabular} \\
\bottomrule
\end{tabular}
\par\vspace{2pt}
{\scriptsize Hidden+Semantic: the per-rotation choices in the canonical semantic-matching files are identical across rotations for every question, so Flip and RStd are 0. Paired contrast CIs: Table~\ref{tab:2x2-contrasts-mmlu}.}
\end{table*}

\begin{table*}[t]
\centering
\footnotesize
\setlength{\tabcolsep}{4pt}
\caption{Two-stage $2\times2$ diagnostic on ARC-Challenge, API models. Rows: option visibility in the answer-generating prompt (hidden/visible) $\times$ matcher (LLM/semantic). Each cell, top to bottom: Accuracy (\%) with 95\% Clopper--Pearson CI; Flip (pp) and RStd (pp) with 95\% question-bootstrap CIs; scored/total where fewer than all questions were scored. Hidden+LLM is the Two-stage condition; Visible+Semantic is the Text-extraction (cascade) condition.}
\label{tab:2x2-arc}
\begin{tabular}{@{}lcccc@{}}
\toprule
Condition & GPT-4.1 mini & Claude Haiku 4.5 & Qwen2.5-7B & Llama-3.1-8B \\
\midrule
Hidden+LLM & \begin{tabular}[t]{@{}c@{}}94.62 [93.17, 95.84] \\ 5.89 [4.61, 7.25] \\ 0.39 [0.18, 0.81]\end{tabular} & \begin{tabular}[t]{@{}c@{}}92.49 [90.83, 93.93] \\ 9.30 [7.68, 10.92] \\ 0.74 [0.33, 1.28] \\ {\scriptsize 1169/1172 scored}\end{tabular} & \begin{tabular}[t]{@{}c@{}}84.04 [81.82, 86.10] \\ 19.97 [17.66, 22.27] \\ 2.64 [1.97, 3.39]\end{tabular} & \begin{tabular}[t]{@{}c@{}}69.97 [67.25, 72.58] \\ 36.26 [33.53, 38.91] \\ 7.54 [6.62, 8.49]\end{tabular} \\
Hidden+Semantic & \begin{tabular}[t]{@{}c@{}}53.41 [50.51, 56.30] \\ 0.00 [0.00, 0.00] \\ 0.00 [0.00, 0.00] \\ {\scriptsize 1042/1172 scored}\end{tabular} & \begin{tabular}[t]{@{}c@{}}50.85 [47.95, 53.75] \\ 0.00 [0.00, 0.00] \\ 0.00 [0.00, 0.00] \\ {\scriptsize 1005/1172 scored}\end{tabular} & \begin{tabular}[t]{@{}c@{}}48.12 [45.23, 51.03] \\ 0.00 [0.00, 0.00] \\ 0.00 [0.00, 0.00] \\ {\scriptsize 1027/1172 scored}\end{tabular} & \begin{tabular}[t]{@{}c@{}}42.83 [39.98, 45.72] \\ 0.00 [0.00, 0.00] \\ 0.00 [0.00, 0.00] \\ {\scriptsize 1002/1172 scored}\end{tabular} \\
Visible+LLM & \begin{tabular}[t]{@{}c@{}}96.08 [94.80, 97.11] \\ 5.89 [4.61, 7.25] \\ 0.44 [0.20, 0.85]\end{tabular} & \begin{tabular}[t]{@{}c@{}}94.28 [92.80, 95.54] \\ 5.74 [4.44, 7.11] \\ 0.40 [0.19, 0.79] \\ {\scriptsize 1167/1172 scored}\end{tabular} & \begin{tabular}[t]{@{}c@{}}89.25 [87.33, 90.96] \\ 14.59 [12.63, 16.64] \\ 1.30 [0.80, 1.94]\end{tabular} & \begin{tabular}[t]{@{}c@{}}77.90 [75.41, 80.25] \\ 32.76 [30.12, 35.49] \\ 3.35 [2.56, 4.23]\end{tabular} \\
Visible+Semantic & \begin{tabular}[t]{@{}c@{}}95.90 [94.61, 96.97] \\ 6.14 [4.78, 7.51] \\ 0.46 [0.20, 0.89]\end{tabular} & \begin{tabular}[t]{@{}c@{}}94.54 [93.08, 95.77] \\ 5.55 [4.27, 6.91] \\ 0.37 [0.19, 0.76] \\ {\scriptsize 1171/1172 scored}\end{tabular} & \begin{tabular}[t]{@{}c@{}}78.84 [76.39, 81.15] \\ 13.24 [11.32, 15.26] \\ 1.51 [0.96, 2.20] \\ {\scriptsize 1035/1172 scored}\end{tabular} & \begin{tabular}[t]{@{}c@{}}75.94 [73.39, 78.36] \\ 34.47 [31.74, 37.20] \\ 3.08 [2.24, 4.04] \\ {\scriptsize 1159/1172 scored}\end{tabular} \\
\bottomrule
\end{tabular}
\par\vspace{2pt}
{\scriptsize Hidden+Semantic: the per-rotation choices in the canonical semantic-matching files are identical across rotations for every question, so Flip and RStd are 0. Paired contrast CIs: Table~\ref{tab:2x2-contrasts-arc}.}
\end{table*}

This does not mean that semantic matching is inherently bad, considering that making the options visible at Stage 1 substantially rescues the accuracy of semantic matching across all models and benchmarks. This is consistent with hidden-option free-text responses being less lexically aligned with the available options, making them harder for the matcher to map reliably to a candidate answer.

The effect of option visibility is smaller and more model-dependent when using the LLM matcher. Visible+LLM usually achieves the highest point-estimate accuracy among the diagnostic conditions, although this is more mixed for Claude. Qwen is also a notable exception under Visible+Semantic, where accuracy remains comparatively poor despite the options being visible. Overall, these results show an interaction between option visibility and matcher type: the effect of the matcher depends strongly on what information is available when the model generates its initial response. Notably, Hidden+Semantic achieves zero Flip by construction despite its sharp accuracy loss, further showing that low order sensitivity does not necessarily imply high accuracy.

Paired contrasts for the \texorpdfstring{$2\times2$}{2x2} diagnostic are reported in Tables~\ref{tab:2x2-contrasts-mmlu} and~\ref{tab:2x2-contrasts-arc} in the Appendix.

\subsection{Cost and Additional Robustness}

Direct and Reasoning Direct use one request per question. Cyclic and Reasoning Cyclic use 4 requests for a four-option question, with IHS also using four calls. Generation-then-matching approaches use one call for Stage 1, then either another call for the second stage, or semantic matching which removes that second call but incurs local matching computation. Reasoning variants have the same request counts as their non-reasoning counterparts, but they usually generate more tokens. With this in mind, it is important that no strategy be judged without taking into account the inference overhead.

Additional robustness analyses, including Holm-corrected exact McNemar tests and repeated-run stochasticity checks, are reported in Appendix~\ref{appendix:results}.

\section{Discussion}

The central finding is that reducing option-order sensitivity does not reliably translate into higher MCQ accuracy. Across the paired accuracy, Flip, and RStd comparisons, we observe several different patterns: in some cases accuracy and order sensitivity improve together, in others both worsen, while in others the three metrics move in different directions. IHS provides one of the clearest examples. Although it is option-order invariant by construction, it improves point-estimate accuracy in only 3/8 model--benchmark cells while decreasing it in the other five. Hidden+Semantic provides an even more extreme case: despite achieving zero Flip by construction, its accuracy remains poor across models and benchmarks. Together, these results suggest that removing or reducing one form of order sensitivity does not necessarily improve the overall quality of MCQ evaluation.

Looking more closely at the order-sensitivity metrics, it is evident that flip rate and RStd are not equivalent. Flip rate asks whether the same question's answer changes under permutation, while RStd measures differences in performance across gold-answer positions. IHS has no access to positional information, yet its RStd can still be nonzero. Of the 41 questions whose gold answer is a relational option, 37 occur at position D; removing those raises GPT's D-position accuracy from 77.4 to 82.6, showing how content composition can create an apparent positional imbalance. This demonstrates that nonzero single-ordering RStd alone is insufficient evidence of positional bias: because it compares different groups of questions, it can also reflect differences in question composition and difficulty across gold-answer positions. Computing RStd over rotations removes this confound, which is what we do for all other RStd computations.

For the generation-then-matching family of approaches, the $2\times2$ diagnostic factorizes different parts of the pipeline and allows us to observe different parts in isolation. The most standout group of cells is Hidden+Semantic, with sharp losses in accuracy for every single cell. However, semantic matching paired with visible options in Stage 1 substantially rescues the accuracy, so semantic matching is not inherently the issue. Instead, the losses arise from combining hidden options with semantic matching, plausibly because free-text answers generated without the options are less lexically aligned with them, making them harder for the semantic matcher to map. We can see this more subtly when using an LLM in the second stage, where having the options visible generally beats hiding them, except for Claude+MMLU where it is effectively unchanged. Taken together, these results demonstrate that withholding the candidate options during generation is not a neutral modification to two-stage evaluation because it can reduce measured accuracy and place greater demands on the downstream matching stage. 

Overall, these results show that MCQ evaluation is brittle in multiple ways, both in how models are asked to answer, and in how robustness is measured. No single metric is sufficient to fully capture MCQ evaluation, and across the models and benchmarks studied, no strategy uniformly dominates across cost, accuracy, and order sensitivity. Evaluations should therefore report accuracy alongside both per-question stability and positional imbalance computed over rotations, while also accounting for unscored outputs where applicable. Robustness interventions should be treated as trade-offs across these dimensions rather than assumed improvements.

\section{Conclusion}

In conclusion, we find that reducing option-order sensitivity does not reliably improve MCQ accuracy. IHS is the clearest example: it is positionally invariant by construction yet does not reliably yield accuracy gains, and its single-ordering RStd can still be non-trivial. Accuracy, flip rate and RStd also capture distinct properties of MCQ evaluation and are not interchangeable. The tested $2\times2$ diagnostic further shows why generation-then-matching approaches can lose accuracy, as withholding the options during Stage 1 may produce outputs that are harder for a semantic matcher to map back to the candidate set.

Across all the tested models and benchmarks studied, no single strategy dominates across cost, accuracy, order sensitivity, and practical failure modes. It is undoubtedly useful to make MCQ evaluation less order-sensitive, but stability should not be mistaken for better performance, and better performance should not be mistaken for more stable answering.

\section*{Limitations}

Our tests only cover four small models on two benchmarks. MMLU is sampled at 20 questions per subject, and almost all questions have four options; a small number of ARC-Challenge questions have a different option count and are retained, with RStd comparisons restricted to four-option questions. Therefore, it is uncertain whether our findings extend to settings with more options (e.g., MMLU-Pro), other question distributions, or larger models.

Although all generation uses temperature 0, hosted APIs are not guaranteed to be deterministic, and our stochasticity checks cover only a 100-question subset per benchmark for four conditions; cyclic permutation and IHS were not repeated, so some observed flips may reflect provider-side non-determinism rather than order sensitivity. For IHS, accuracy is also sensitive to deterministic tie-breaking, particularly for Qwen and Llama, where a substantial proportion of questions have tied top scores. Flip rate and RStd are computed over parsed outputs only, and conditions with substantial unscored outputs, most notably Hidden+Semantic and Qwen Visible+Semantic, may have unequal effective question sets across positions, which may affect the interpretation of the values and how they compare against other RStds.

Each strategy was tested with only a single set of prompt templates; no prompt ablations were tested. Additionally, the LLM matcher is always the same model as the generator. We suspect that the $2\times2$ result where Hidden+Semantic falls sharply in accuracy is because the model may produce a lexically misaligned Stage 1 output with the candidate options, which is difficult for the semantic matcher to match correctly.

Finally, Llama was served through two DeepInfra-hosted routes: one for cyclic permutation, Reasoning Cyclic and IHS (from which Direct and Reasoning Direct are also derived), and another for the two-stage and $2\times2$ conditions. Both routes were documented as serving FP8 versions of Llama-3.1-8B-Instruct with identical decoding settings, but we could not verify that their underlying deployments were equivalent. Consequently, Direct-relative comparisons involving the two-stage family may be confounded by deployment differences. The serving precision of the Qwen deployment is also undocumented.

\section*{Ethical Considerations}
Generative AI tools were used to assist with both the codebase and manuscript. In the codebase, AI assistance was used for feature implementation and project-structure improvements. In the manuscript, AI assistance was used for wording refinement, LaTeX formatting, table presentation, and claim clarification. The authors remain responsible for the experimental design, implementation, analysis, interpretation, and final claims.

This is an evaluation-methodology study using public benchmarks with no human subjects or new data collection, so risks are minimal; the main consideration is that debiasing methods reported as ineffective should not be relied upon as safety guarantees.

\bibliography{refs}
\appendix
\raggedbottom
\section{Full Accuracy Results}
\label{appendix:accuracy}

Tables~\ref{tab:accuracy-mmlu} and~\ref{tab:accuracy-arc} report end-to-end (E2E) accuracy for all ten conditions and the four API models on MMLU ($n=1140$) and ARC-Challenge ($n=1172$). Unscored questions count as incorrect on E2E, and 95\% confidence intervals are exact Clopper--Pearson intervals. Where fewer than all questions were scored, the number scored is shown beneath the estimate. Tables~\ref{tab:conditional-mmlu} and~\ref{tab:conditional-arc} contrast E2E accuracy with conditional accuracy (correct / scored), each with a Clopper--Pearson interval over its own denominator, for every condition with unscored questions.

\begin{table*}[t]
\centering
\footnotesize
\setlength{\tabcolsep}{4pt}
\caption{End-to-end accuracy (\%) on MMLU ($n=1140$), API models, with 95\% Clopper--Pearson confidence intervals. Unscored questions count as incorrect; where fewer than all questions were scored, the number scored is shown beneath the estimate.}
\label{tab:accuracy-mmlu}
\begin{tabular}{@{}lcccc@{}}
\toprule
Method & GPT-4.1 mini & Claude Haiku 4.5 & Qwen2.5-7B & Llama-3.1-8B \\
\midrule
Direct & \begin{tabular}[t]{@{}c@{}}81.84 [79.48, 84.04] \\ {\scriptsize 1139/1140 scored}\end{tabular} & \begin{tabular}[t]{@{}c@{}}84.47 [82.24, 86.53] \\ {\scriptsize 1105/1140 scored}\end{tabular} & 74.30 [71.66, 76.81] & \begin{tabular}[t]{@{}c@{}}59.12 [56.20, 61.99] \\ {\scriptsize 1139/1140 scored}\end{tabular} \\
Reasoning Direct & \begin{tabular}[t]{@{}c@{}}89.12 [87.17, 90.87] \\ {\scriptsize 1137/1140 scored}\end{tabular} & \begin{tabular}[t]{@{}c@{}}86.75 [84.65, 88.67] \\ {\scriptsize 1134/1140 scored}\end{tabular} & \begin{tabular}[t]{@{}c@{}}75.26 [72.65, 77.74] \\ {\scriptsize 1139/1140 scored}\end{tabular} & \begin{tabular}[t]{@{}c@{}}65.61 [62.78, 68.37] \\ {\scriptsize 1133/1140 scored}\end{tabular} \\
Cyclic & 82.11 [79.75, 84.29] & \begin{tabular}[t]{@{}c@{}}85.44 [83.26, 87.44] \\ {\scriptsize 1127/1140 scored}\end{tabular} & 75.18 [72.56, 77.66] & 61.93 [59.04, 64.76] \\
Reasoning Cyclic & 88.95 [86.98, 90.71] & \begin{tabular}[t]{@{}c@{}}90.26 [88.39, 91.92] \\ {\scriptsize 1139/1140 scored}\end{tabular} & 78.42 [75.92, 80.78] & 70.70 [67.96, 73.33] \\
Two-stage & \begin{tabular}[t]{@{}c@{}}81.58 [79.20, 83.79] \\ {\scriptsize 1139/1140 scored}\end{tabular} & \begin{tabular}[t]{@{}c@{}}83.07 [80.76, 85.20] \\ {\scriptsize 1128/1140 scored}\end{tabular} & \begin{tabular}[t]{@{}c@{}}68.51 [65.72, 71.20] \\ {\scriptsize 1139/1140 scored}\end{tabular} & \begin{tabular}[t]{@{}c@{}}59.39 [56.47, 62.25] \\ {\scriptsize 1136/1140 scored}\end{tabular} \\
Reasoning Two-stage & 85.79 [83.63, 87.77] & \begin{tabular}[t]{@{}c@{}}86.75 [84.65, 88.67] \\ {\scriptsize 1126/1140 scored}\end{tabular} & 70.35 [67.61, 72.99] & \begin{tabular}[t]{@{}c@{}}59.65 [56.74, 62.51] \\ {\scriptsize 1139/1140 scored}\end{tabular} \\
IHS & \begin{tabular}[t]{@{}c@{}}82.54 [80.21, 84.71] \\ {\scriptsize 1139/1140 scored}\end{tabular} & 85.88 [83.72, 87.85] & 67.54 [64.74, 70.26] & \begin{tabular}[t]{@{}c@{}}55.88 [52.94, 58.79] \\ {\scriptsize 1137/1140 scored}\end{tabular} \\
Hidden+Semantic & \begin{tabular}[t]{@{}c@{}}46.67 [43.74, 49.61] \\ {\scriptsize 985/1140 scored}\end{tabular} & \begin{tabular}[t]{@{}c@{}}44.74 [41.82, 47.68] \\ {\scriptsize 910/1140 scored}\end{tabular} & \begin{tabular}[t]{@{}c@{}}40.70 [37.83, 43.62] \\ {\scriptsize 973/1140 scored}\end{tabular} & \begin{tabular}[t]{@{}c@{}}37.37 [34.55, 40.25] \\ {\scriptsize 941/1140 scored}\end{tabular} \\
Visible+LLM & 84.21 [81.96, 86.28] & \begin{tabular}[t]{@{}c@{}}82.98 [80.67, 85.12] \\ {\scriptsize 1111/1140 scored}\end{tabular} & 73.25 [70.57, 75.80] & \begin{tabular}[t]{@{}c@{}}63.07 [60.19, 65.88] \\ {\scriptsize 1139/1140 scored}\end{tabular} \\
Visible+Semantic & \begin{tabular}[t]{@{}c@{}}83.42 [81.13, 85.54] \\ {\scriptsize 1138/1140 scored}\end{tabular} & \begin{tabular}[t]{@{}c@{}}86.23 [84.09, 88.18] \\ {\scriptsize 1133/1140 scored}\end{tabular} & \begin{tabular}[t]{@{}c@{}}59.21 [56.29, 62.08] \\ {\scriptsize 921/1140 scored}\end{tabular} & \begin{tabular}[t]{@{}c@{}}59.04 [56.12, 61.91] \\ {\scriptsize 1086/1140 scored}\end{tabular} \\
\bottomrule
\end{tabular}
\end{table*}

\begin{table*}[t]
\centering
\footnotesize
\setlength{\tabcolsep}{4pt}
\caption{End-to-end accuracy (\%) on ARC-Challenge ($n=1172$), API models, with 95\% Clopper--Pearson confidence intervals. Unscored questions count as incorrect; where fewer than all questions were scored, the number scored is shown beneath the estimate.}
\label{tab:accuracy-arc}
\begin{tabular}{@{}lcccc@{}}
\toprule
Method & GPT-4.1 mini & Claude Haiku 4.5 & Qwen2.5-7B & Llama-3.1-8B \\
\midrule
Direct & 95.82 [94.51, 96.89] & \begin{tabular}[t]{@{}c@{}}94.54 [93.08, 95.77] \\ {\scriptsize 1168/1172 scored}\end{tabular} & 89.42 [87.52, 91.12] & 77.22 [74.71, 79.59] \\
Reasoning Direct & \begin{tabular}[t]{@{}c@{}}96.16 [94.90, 97.19] \\ {\scriptsize 1170/1172 scored}\end{tabular} & \begin{tabular}[t]{@{}c@{}}94.28 [92.80, 95.54] \\ {\scriptsize 1171/1172 scored}\end{tabular} & \begin{tabular}[t]{@{}c@{}}90.96 [89.17, 92.54] \\ {\scriptsize 1171/1172 scored}\end{tabular} & \begin{tabular}[t]{@{}c@{}}83.62 [81.37, 85.69] \\ {\scriptsize 1171/1172 scored}\end{tabular} \\
Cyclic & 96.16 [94.90, 97.19] & 95.56 [94.22, 96.67] & 90.87 [89.07, 92.46] & 80.63 [78.25, 82.86] \\
Reasoning Cyclic & 97.27 [96.17, 98.13] & 96.33 [95.09, 97.33] & 93.00 [91.39, 94.40] & 88.40 [86.42, 90.17] \\
Two-stage & 94.62 [93.17, 95.84] & \begin{tabular}[t]{@{}c@{}}92.49 [90.83, 93.93] \\ {\scriptsize 1169/1172 scored}\end{tabular} & 84.04 [81.82, 86.10] & 69.97 [67.25, 72.58] \\
Reasoning Two-stage & \begin{tabular}[t]{@{}c@{}}94.28 [92.80, 95.54] \\ {\scriptsize 1170/1172 scored}\end{tabular} & \begin{tabular}[t]{@{}c@{}}92.15 [90.46, 93.63] \\ {\scriptsize 1164/1172 scored}\end{tabular} & 80.72 [78.34, 82.94] & 73.63 [71.01, 76.14] \\
IHS & 93.86 [92.33, 95.16] & 95.05 [93.65, 96.22] & 84.13 [81.91, 86.18] & 71.50 [68.82, 74.07] \\
Hidden+Semantic & \begin{tabular}[t]{@{}c@{}}53.41 [50.51, 56.30] \\ {\scriptsize 1042/1172 scored}\end{tabular} & \begin{tabular}[t]{@{}c@{}}50.85 [47.95, 53.75] \\ {\scriptsize 1005/1172 scored}\end{tabular} & \begin{tabular}[t]{@{}c@{}}48.12 [45.23, 51.03] \\ {\scriptsize 1027/1172 scored}\end{tabular} & \begin{tabular}[t]{@{}c@{}}42.83 [39.98, 45.72] \\ {\scriptsize 1002/1172 scored}\end{tabular} \\
Visible+LLM & 96.08 [94.80, 97.11] & \begin{tabular}[t]{@{}c@{}}94.28 [92.80, 95.54] \\ {\scriptsize 1167/1172 scored}\end{tabular} & 89.25 [87.33, 90.96] & 77.90 [75.41, 80.25] \\
Visible+Semantic & 95.90 [94.61, 96.97] & \begin{tabular}[t]{@{}c@{}}94.54 [93.08, 95.77] \\ {\scriptsize 1171/1172 scored}\end{tabular} & \begin{tabular}[t]{@{}c@{}}78.84 [76.39, 81.15] \\ {\scriptsize 1035/1172 scored}\end{tabular} & \begin{tabular}[t]{@{}c@{}}75.94 [73.39, 78.36] \\ {\scriptsize 1159/1172 scored}\end{tabular} \\
\bottomrule
\end{tabular}
\end{table*}

\begin{table*}[t]
\centering
\footnotesize
\setlength{\tabcolsep}{4pt}
\caption{End-to-end versus conditional accuracy on MMLU, API models. End-to-end: correct / all 1140 questions (unscored count as incorrect). Conditional: correct / scored questions. 95\% Clopper--Pearson CIs with the respective denominator. Gap: conditional minus end-to-end, in percentage points.}
\label{tab:conditional-mmlu}
\begin{tabular}{@{}lcccr@{}}
\toprule
Condition & End-to-end [95\% CI] & Conditional [95\% CI] & Scored/total & Gap \\
\midrule
\multicolumn{5}{@{}l}{\textit{GPT-4.1 mini}} \\
\quad Direct & 81.84 [79.48, 84.04] & 81.91 [79.55, 84.11] & 1139/1140 & 0.07 \\
\quad Reasoning Direct & 89.12 [87.17, 90.87] & 89.36 [87.42, 91.09] & 1137/1140 & 0.24 \\
\quad Two-stage & 81.58 [79.20, 83.79] & 81.65 [79.28, 83.86] & 1139/1140 & 0.07 \\
\quad IHS & 82.54 [80.21, 84.71] & 82.62 [80.29, 84.77] & 1139/1140 & 0.07 \\
\quad Hidden+Semantic & 46.67 [43.74, 49.61] & 54.01 [50.84, 57.16] & 985/1140 & 7.34 \\
\quad Visible+Semantic & 83.42 [81.13, 85.54] & 83.57 [81.28, 85.68] & 1138/1140 & 0.15 \\
\multicolumn{5}{@{}l}{\quad\scriptsize All scored (1140/1140): Cyclic, Reasoning Cyclic, Reasoning Two-stage, Visible+LLM} \\
\multicolumn{5}{@{}l}{\textit{Claude Haiku 4.5}} \\
\quad Direct & 84.47 [82.24, 86.53] & 87.15 [85.03, 89.07] & 1105/1140 & 2.68 \\
\quad Reasoning Direct & 86.75 [84.65, 88.67] & 87.21 [85.13, 89.10] & 1134/1140 & 0.46 \\
\quad Cyclic & 85.44 [83.26, 87.44] & 86.42 [84.28, 88.37] & 1127/1140 & 0.99 \\
\quad Reasoning Cyclic & 90.26 [88.39, 91.92] & 90.34 [88.48, 92.00] & 1139/1140 & 0.08 \\
\quad Two-stage & 83.07 [80.76, 85.20] & 83.95 [81.68, 86.05] & 1128/1140 & 0.88 \\
\quad Reasoning Two-stage & 86.75 [84.65, 88.67] & 87.83 [85.78, 89.69] & 1126/1140 & 1.08 \\
\quad Hidden+Semantic & 44.74 [41.82, 47.68] & 56.04 [52.75, 59.30] & 910/1140 & 11.31 \\
\quad Visible+LLM & 82.98 [80.67, 85.12] & 85.15 [82.92, 87.19] & 1111/1140 & 2.17 \\
\quad Visible+Semantic & 86.23 [84.09, 88.18] & 86.76 [84.65, 88.68] & 1133/1140 & 0.53 \\
\multicolumn{5}{@{}l}{\quad\scriptsize All scored (1140/1140): IHS} \\
\multicolumn{5}{@{}l}{\textit{Qwen2.5-7B}} \\
\quad Reasoning Direct & 75.26 [72.65, 77.74] & 75.33 [72.72, 77.81] & 1139/1140 & 0.07 \\
\quad Two-stage & 68.51 [65.72, 71.20] & 68.57 [65.78, 71.26] & 1139/1140 & 0.06 \\
\quad Hidden+Semantic & 40.70 [37.83, 43.62] & 47.69 [44.51, 50.88] & 973/1140 & 6.99 \\
\quad Visible+Semantic & 59.21 [56.29, 62.08] & 73.29 [70.31, 76.12] & 921/1140 & 14.08 \\
\multicolumn{5}{@{}l}{\quad\scriptsize All scored (1140/1140): Direct, Cyclic, Reasoning Cyclic, Reasoning Two-stage, IHS, Visible+LLM} \\
\multicolumn{5}{@{}l}{\textit{Llama-3.1-8B}} \\
\quad Direct & 59.12 [56.20, 61.99] & 59.17 [56.26, 62.05] & 1139/1140 & 0.05 \\
\quad Reasoning Direct & 65.61 [62.78, 68.37] & 66.02 [63.18, 68.78] & 1133/1140 & 0.41 \\
\quad Two-stage & 59.39 [56.47, 62.25] & 59.60 [56.68, 62.46] & 1136/1140 & 0.21 \\
\quad Reasoning Two-stage & 59.65 [56.74, 62.51] & 59.70 [56.79, 62.57] & 1139/1140 & 0.05 \\
\quad IHS & 55.88 [52.94, 58.79] & 56.02 [53.08, 58.93] & 1137/1140 & 0.15 \\
\quad Hidden+Semantic & 37.37 [34.55, 40.25] & 45.27 [42.06, 48.52] & 941/1140 & 7.90 \\
\quad Visible+LLM & 63.07 [60.19, 65.88] & 63.13 [60.25, 65.93] & 1139/1140 & 0.06 \\
\quad Visible+Semantic & 59.04 [56.12, 61.91] & 61.97 [59.01, 64.87] & 1086/1140 & 2.94 \\
\multicolumn{5}{@{}l}{\quad\scriptsize All scored (1140/1140): Cyclic, Reasoning Cyclic} \\
\bottomrule
\end{tabular}
\par\vspace{2pt}
{\scriptsize Conditions in which every question was scored (conditional = end-to-end) are listed per model without values.}
\end{table*}

\begin{table*}[t]
\centering
\footnotesize
\setlength{\tabcolsep}{4pt}
\caption{End-to-end versus conditional accuracy on ARC-Challenge, API models. End-to-end: correct / all 1172 questions (unscored count as incorrect). Conditional: correct / scored questions. 95\% Clopper--Pearson CIs with the respective denominator. Gap: conditional minus end-to-end, in percentage points.}
\label{tab:conditional-arc}
\begin{tabular}{@{}lcccr@{}}
\toprule
Condition & End-to-end [95\% CI] & Conditional [95\% CI] & Scored/total & Gap \\
\midrule
\multicolumn{5}{@{}l}{\textit{GPT-4.1 mini}} \\
\quad Reasoning Direct & 96.16 [94.90, 97.19] & 96.32 [95.08, 97.33] & 1170/1172 & 0.16 \\
\quad Reasoning Two-stage & 94.28 [92.80, 95.54] & 94.44 [92.97, 95.69] & 1170/1172 & 0.16 \\
\quad Hidden+Semantic & 53.41 [50.51, 56.30] & 60.08 [57.03, 63.07] & 1042/1172 & 6.66 \\
\multicolumn{5}{@{}l}{\quad\scriptsize All scored (1172/1172): Direct, Cyclic, Reasoning Cyclic, Two-stage, IHS, Visible+LLM, Visible+Semantic} \\
\multicolumn{5}{@{}l}{\textit{Claude Haiku 4.5}} \\
\quad Direct & 94.54 [93.08, 95.77] & 94.86 [93.44, 96.06] & 1168/1172 & 0.32 \\
\quad Reasoning Direct & 94.28 [92.80, 95.54] & 94.36 [92.88, 95.61] & 1171/1172 & 0.08 \\
\quad Two-stage & 92.49 [90.83, 93.93] & 92.73 [91.09, 94.15] & 1169/1172 & 0.24 \\
\quad Reasoning Two-stage & 92.15 [90.46, 93.63] & 92.78 [91.14, 94.20] & 1164/1172 & 0.63 \\
\quad Hidden+Semantic & 50.85 [47.95, 53.75] & 59.30 [56.19, 62.36] & 1005/1172 & 8.45 \\
\quad Visible+LLM & 94.28 [92.80, 95.54] & 94.69 [93.24, 95.90] & 1167/1172 & 0.40 \\
\quad Visible+Semantic & 94.54 [93.08, 95.77] & 94.62 [93.17, 95.84] & 1171/1172 & 0.08 \\
\multicolumn{5}{@{}l}{\quad\scriptsize All scored (1172/1172): Cyclic, Reasoning Cyclic, IHS} \\
\multicolumn{5}{@{}l}{\textit{Qwen2.5-7B}} \\
\quad Reasoning Direct & 90.96 [89.17, 92.54] & 91.03 [89.25, 92.61] & 1171/1172 & 0.08 \\
\quad Hidden+Semantic & 48.12 [45.23, 51.03] & 54.92 [51.81, 57.99] & 1027/1172 & 6.79 \\
\quad Visible+Semantic & 78.84 [76.39, 81.15] & 89.28 [87.23, 91.09] & 1035/1172 & 10.44 \\
\multicolumn{5}{@{}l}{\quad\scriptsize All scored (1172/1172): Direct, Cyclic, Reasoning Cyclic, Two-stage, Reasoning Two-stage, IHS, Visible+LLM} \\
\multicolumn{5}{@{}l}{\textit{Llama-3.1-8B}} \\
\quad Reasoning Direct & 83.62 [81.37, 85.69] & 83.69 [81.45, 85.76] & 1171/1172 & 0.07 \\
\quad Hidden+Semantic & 42.83 [39.98, 45.72] & 50.10 [46.96, 53.24] & 1002/1172 & 7.27 \\
\quad Visible+Semantic & 75.94 [73.39, 78.36] & 76.79 [74.25, 79.19] & 1159/1172 & 0.85 \\
\multicolumn{5}{@{}l}{\quad\scriptsize All scored (1172/1172): Direct, Cyclic, Reasoning Cyclic, Two-stage, Reasoning Two-stage, IHS, Visible+LLM} \\
\bottomrule
\end{tabular}
\par\vspace{2pt}
{\scriptsize Conditions in which every question was scored (conditional = end-to-end) are listed per model without values.}
\end{table*}

\section{IHS Score Analysis}
\label{appendix:ihs}

IHS issues one call per option. Each prompt contains the question and a single option phrased as a hypothesis, without option letters and without the other options (Appendix~\ref{appendix:prompts}). The model returns a 0--100 confidence score in \texttt{<score>} tags. A score is valid only if the tag is present and its value lies in $[0,100]$. Invalid scores are excluded from the argmax, and a question with no valid score is unscored. The prediction is the option with the highest score. Table~\ref{tab:ihs-score-analysis} summarizes the per-option scores stored in the canonical IHS outputs.

\paragraph{Tie-breaking.} Exact ties for the maximum score are resolved by a deterministic function: a hash of the experiment seed, benchmark, question ID, method name and the set of tied canonical option IDs. Displayed letters, rotation index and execution order are not inputs. The same utility resolves majority-vote ties in cyclic permutation and ties in the semantic-matching cascade. Score ties therefore do not introduce positional bias. In the stored outputs the prediction is always a member of the tied set and is distributed across first, middle and last tied positions.

\paragraph{Tie-dependent correctness.} Ties occur on 6.2--14.9\% of MMLU questions for GPT-4.1 mini and Claude Haiku 4.5, and on 22.7--26.7\% for Qwen2.5-7B and Llama-3.1-8B (ARC-Challenge: 3.1--6.1\% and 12.0--18.2\%). A question is tie-dependent when the gold option is among the tied top-scoring options, so its correctness depends on the tie-break. Accuracy on tied questions is 31.0--55.6\%, compared with 64.3--97.1\% on untied questions. The choice of tie resolution can therefore affect accuracy: across all possible resolutions, Qwen2.5-7B on MMLU ranges from 59.56\% to 78.33\%, while Llama-3.1-8B ranges from 47.02\% to 68.25\%. For Qwen2.5-7B and Llama-3.1-8B, roughly one third of errors occur when the gold option is tied for the top score but loses the tie-break; the remaining errors occur when the gold option is strictly outranked.

\paragraph{Score granularity.} Models differ substantially in how precisely they report IHS confidence scores. At least 99\% of valid scores from GPT-4.1 mini, Qwen2.5-7B and Llama-3.1-8B are multiples of 5; for Llama-3.1-8B, 90.6--91.4\% are multiples of 10. Claude Haiku 4.5 produces finer scores, with only 32.9--39.3\% being multiples of 5 and 0.5--1.0\% multiples of 10. Claude also has the lowest tie rates.

\paragraph{Margins.} The top-1 minus top-2 margin is reported on the 0--100 scale (Table~\ref{tab:ihs-score-analysis}). Median margins range from 13 (Llama-3.1-8B, MMLU) to 85 (GPT-4.1 mini, ARC-Challenge). For Llama-3.1-8B on MMLU, the median difference between the gold option's score and the strongest distractor's score is 0.

\paragraph{Relational options.} An item counts as containing a relational option if any option matches an ``all/none/both/neither of the above''-type pattern. This applies to 78 MMLU items (6.8\%) and 15 ARC-Challenge items. Because IHS scores each option in isolation, these options are judged without the options they refer to. On the 78 MMLU items, IHS accuracy is 53.8\% (GPT-4.1 mini), 60.3\% (Claude Haiku 4.5), 44.9\% (Qwen2.5-7B) and 46.2\% (Llama-3.1-8B), compared with 75.6\%, 83.3\%, 64.1\% and 57.7\% for Direct. On the 15 ARC-Challenge items the two coincide for GPT-4.1 mini (100.0\%), Claude Haiku 4.5 (100.0\%) and Qwen2.5-7B (93.3\%); for Llama-3.1-8B they are 46.7\% (IHS) and 73.3\% (Direct).

\paragraph{Position D and IHS RStd.} IHS presents no option positions, so no Flip is measured, and no rotated IHS runs were performed. Its RStd is therefore computed from the single canonical ordering: the standard deviation of accuracy across gold-answer positions, which compares different groups of questions. These values are 3.20 (GPT-4.1 mini), 3.47 (Claude Haiku 4.5), 3.55 (Qwen2.5-7B) and 2.26 (Llama-3.1-8B) on MMLU, and 1.32, 0.78, 3.59 and 2.32 on ARC-Challenge. On MMLU, 41 items have a relational gold option, and 37 of these place it at position D (A: 0, B: 1, C: 3). For GPT-4.1 mini, IHS accuracy at position D is 77.4\% overall and 82.6\% after excluding gold-relational items, against 83.5--86.6\% at positions A--C. Because they come from a single ordering, these RStd values are not directly comparable to the rotation-based RStd in Tables~\ref{tab:order-mmlu} and~\ref{tab:order-arc}.

\begin{table*}[t]
\centering
\footnotesize
\setlength{\tabcolsep}{4pt}
\caption{Score-level analysis of IHS, API models. IHS accuracy is end-to-end (\%). Tie: multiple options share the maximum score. Tie-dependent: the gold option is among the tied top options, so tie-breaking determines correctness. Gold unique-top: the gold option alone has the maximum score. Margin: top-1 minus top-2 score on the 0--100 scale (median [interquartile range]). Percentages are of all questions ($n=1140$ MMLU, $n=1172$ ARC-Challenge).}
\label{tab:ihs-score-analysis}
\begin{tabular}{@{}llrrrrc@{}}
\toprule
Model & Benchmark & IHS Acc. & Tie \% & Tie-dep.\ \% & Gold unique-top \% & Margin [IQR] \\
\midrule
GPT-4.1 mini     & MMLU & 82.54 & 14.9 & 12.7 & 77.4 & 75 [5, 85] \\
                 & ARC  & 93.86 &  6.1 &  5.7 & 91.2 & 85 [55, 90] \\
Claude Haiku 4.5 & MMLU & 85.88 &  6.2 &  4.9 & 83.9 & 67 [27, 87] \\
                 & ARC  & 95.05 &  3.1 &  2.8 & 93.3 & 67 [37, 84] \\
Qwen2.5-7B       & MMLU & 67.54 & 22.7 & 18.8 & 59.6 & 25 [5, 80] \\
                 & ARC  & 84.13 & 12.0 & 11.2 & 80.0 & 55 [15, 85] \\
Llama-3.1-8B     & MMLU & 55.88 & 26.7 & 21.2 & 47.0 & 13 [0, 50] \\
                 & ARC  & 71.50 & 18.2 & 15.4 & 65.5 & 30 [5, 70] \\
\bottomrule
\end{tabular}
\par\vspace{2pt}
{\scriptsize Ties are resolved deterministically from canonical option IDs, not displayed positions, so they do not introduce positional bias.}
\end{table*}

\section{Full Order-Sensitivity Results}
\label{appendix:order}

Tables~\ref{tab:order-mmlu} and~\ref{tab:order-arc} report Flip and RStd for every condition with per-rotation outputs. Flip is the percentage of questions whose parsed choice, mapped back to the original option order, is not identical across all cyclic rotations of the options. It is computed over questions with at least one parsed rotation. RStd is the standard deviation, across answer positions, of position-specific recall (recall by position of the correct option), in percentage points pooled over all cyclic rotations. Direct and Reasoning Direct use the rotations of the cyclic permutation and Reasoning Cyclic runs, whose rotation~0 is the Direct observation. For Two-stage, Reasoning Two-stage and Visible+LLM, rotations~1 to $n-1$ re-issue only the matching request. The stage-1 text is reused: for Two-stage and Reasoning Two-stage, the saved hidden-option answer; for Visible+LLM, the text-extraction output of the same rotation. Visible+Semantic re-issues the full request under each rotation. Rotation~0 is always the canonical production output. 95\% CIs come from a question-level percentile bootstrap with 10{,}000 resamples, recomputing each metric on every resample.

\begin{table*}[t]
\centering
\footnotesize
\setlength{\tabcolsep}{4pt}
\caption{Order sensitivity on MMLU, API models. Each cell: Flip [95\% CI] (top) / RStd [95\% CI] (bottom). Flip: share of questions whose parsed choice (mapped to the original option order) is not identical across all cyclic option rotations, in percentage points, over questions with at least one parsed rotation. RStd: standard deviation of position-specific recall (recall by position of the correct option), in percentage points, over 4-option questions. Direct and Reasoning Direct use the rotations of the Cyclic and Reasoning Cyclic runs. 95\% CIs: question-level percentile bootstrap (10{,}000 resamples).}
\label{tab:order-mmlu}
\begin{tabular}{@{}lcccc@{}}
\toprule
Method & GPT-4.1 mini & Claude Haiku 4.5 & Qwen2.5-7B & Llama-3.1-8B \\
\midrule
Direct & \begin{tabular}[t]{@{}c@{}}20.26 [17.98, 22.63] \\ 2.02 [1.41, 2.73]\end{tabular} & \begin{tabular}[t]{@{}c@{}}22.54 [20.14, 25.02] \\ 0.42 [0.23, 1.31]\end{tabular} & \begin{tabular}[t]{@{}c@{}}31.23 [28.51, 33.95] \\ 2.29 [1.61, 3.12]\end{tabular} & \begin{tabular}[t]{@{}c@{}}54.74 [51.84, 57.72] \\ 5.56 [4.58, 6.65]\end{tabular} \\
Reasoning Direct & \begin{tabular}[t]{@{}c@{}}13.07 [11.14, 15.09] \\ 0.61 [0.24, 1.27]\end{tabular} & \begin{tabular}[t]{@{}c@{}}21.25 [18.89, 23.62] \\ 0.68 [0.31, 1.46]\end{tabular} & \begin{tabular}[t]{@{}c@{}}33.77 [31.05, 36.58] \\ 3.59 [2.77, 4.50]\end{tabular} & \begin{tabular}[t]{@{}c@{}}53.25 [50.35, 56.14] \\ 0.81 [0.37, 2.03]\end{tabular} \\
Two-stage & \begin{tabular}[t]{@{}c@{}}12.63 [10.70, 14.56] \\ 0.45 [0.20, 1.06]\end{tabular} & \begin{tabular}[t]{@{}c@{}}16.51 [14.39, 18.61] \\ 1.45 [0.91, 2.13]\end{tabular} & \begin{tabular}[t]{@{}c@{}}31.75 [29.12, 34.47] \\ 3.22 [2.42, 4.11]\end{tabular} & \begin{tabular}[t]{@{}c@{}}49.38 [46.44, 52.28] \\ 8.81 [7.76, 9.91]\end{tabular} \\
Reasoning Two-stage & \begin{tabular}[t]{@{}c@{}}10.96 [9.12, 12.81] \\ 0.78 [0.40, 1.32]\end{tabular} & \begin{tabular}[t]{@{}c@{}}14.85 [12.82, 16.89] \\ 1.88 [1.32, 2.53]\end{tabular} & \begin{tabular}[t]{@{}c@{}}31.75 [29.12, 34.47] \\ 3.35 [2.47, 4.32]\end{tabular} & \begin{tabular}[t]{@{}c@{}}47.06 [44.21, 49.91] \\ 9.38 [8.28, 10.52]\end{tabular} \\
Visible+LLM & \begin{tabular}[t]{@{}c@{}}16.75 [14.65, 18.95] \\ 1.29 [0.79, 1.94]\end{tabular} & \begin{tabular}[t]{@{}c@{}}20.59 [18.25, 22.95] \\ 0.92 [0.40, 1.72]\end{tabular} & \begin{tabular}[t]{@{}c@{}}31.84 [29.12, 34.56] \\ 2.51 [1.81, 3.35]\end{tabular} & \begin{tabular}[t]{@{}c@{}}53.77 [50.88, 56.67] \\ 4.25 [3.28, 5.35]\end{tabular} \\
Visible+Semantic & \begin{tabular}[t]{@{}c@{}}16.93 [14.82, 19.21] \\ 1.28 [0.78, 1.93]\end{tabular} & \begin{tabular}[t]{@{}c@{}}16.04 [13.92, 18.15] \\ 0.85 [0.43, 1.50]\end{tabular} & \begin{tabular}[t]{@{}c@{}}29.36 [26.60, 32.12] \\ 2.14 [1.35, 3.15]\end{tabular} & \begin{tabular}[t]{@{}c@{}}51.58 [48.59, 54.53] \\ 4.10 [3.14, 5.21]\end{tabular} \\
\bottomrule
\end{tabular}
\end{table*}

\begin{table*}[t]
\centering
\footnotesize
\setlength{\tabcolsep}{4pt}
\caption{Order sensitivity on ARC-Challenge, API models. Each cell: Flip [95\% CI] (top) / RStd [95\% CI] (bottom). Flip: share of questions whose parsed choice (mapped to the original option order) is not identical across all cyclic option rotations, in percentage points, over questions with at least one parsed rotation. RStd: standard deviation of position-specific recall (recall by position of the correct option), in percentage points, over 4-option questions. Direct and Reasoning Direct use the rotations of the Cyclic and Reasoning Cyclic runs. 95\% CIs: question-level percentile bootstrap (10{,}000 resamples).}
\label{tab:order-arc}
\begin{tabular}{@{}lcccc@{}}
\toprule
Method & GPT-4.1 mini & Claude Haiku 4.5 & Qwen2.5-7B & Llama-3.1-8B \\
\midrule
Direct & \begin{tabular}[t]{@{}c@{}}6.06 [4.69, 7.42] \\ 0.18 [0.10, 0.66]\end{tabular} & \begin{tabular}[t]{@{}c@{}}8.36 [6.83, 9.98] \\ 0.67 [0.30, 1.20]\end{tabular} & \begin{tabular}[t]{@{}c@{}}14.16 [12.20, 16.13] \\ 0.89 [0.46, 1.52]\end{tabular} & \begin{tabular}[t]{@{}c@{}}29.95 [27.30, 32.59] \\ 2.26 [1.55, 3.08]\end{tabular} \\
Reasoning Direct & \begin{tabular}[t]{@{}c@{}}3.75 [2.73, 4.86] \\ 0.13 [0.07, 0.49]\end{tabular} & \begin{tabular}[t]{@{}c@{}}11.01 [9.30, 12.80] \\ 0.62 [0.30, 1.17]\end{tabular} & \begin{tabular}[t]{@{}c@{}}14.85 [12.80, 16.81] \\ 2.12 [1.53, 2.80]\end{tabular} & \begin{tabular}[t]{@{}c@{}}31.23 [28.58, 33.87] \\ 1.36 [0.72, 2.29]\end{tabular} \\
Two-stage & \begin{tabular}[t]{@{}c@{}}5.89 [4.61, 7.25] \\ 0.39 [0.18, 0.81]\end{tabular} & \begin{tabular}[t]{@{}c@{}}9.30 [7.68, 10.92] \\ 0.74 [0.33, 1.28]\end{tabular} & \begin{tabular}[t]{@{}c@{}}19.97 [17.66, 22.27] \\ 2.64 [1.97, 3.39]\end{tabular} & \begin{tabular}[t]{@{}c@{}}36.26 [33.53, 38.91] \\ 7.54 [6.62, 8.49]\end{tabular} \\
Reasoning Two-stage & \begin{tabular}[t]{@{}c@{}}5.38 [4.10, 6.66] \\ 0.47 [0.25, 0.84]\end{tabular} & \begin{tabular}[t]{@{}c@{}}9.48 [7.85, 11.18] \\ 0.94 [0.50, 1.49]\end{tabular} & \begin{tabular}[t]{@{}c@{}}26.62 [24.06, 29.18] \\ 5.39 [4.53, 6.31]\end{tabular} & \begin{tabular}[t]{@{}c@{}}36.18 [33.45, 38.91] \\ 8.85 [7.88, 9.88]\end{tabular} \\
Visible+LLM & \begin{tabular}[t]{@{}c@{}}5.89 [4.61, 7.25] \\ 0.44 [0.20, 0.85]\end{tabular} & \begin{tabular}[t]{@{}c@{}}5.74 [4.44, 7.11] \\ 0.40 [0.19, 0.79]\end{tabular} & \begin{tabular}[t]{@{}c@{}}14.59 [12.63, 16.64] \\ 1.30 [0.80, 1.94]\end{tabular} & \begin{tabular}[t]{@{}c@{}}32.76 [30.12, 35.49] \\ 3.35 [2.56, 4.23]\end{tabular} \\
Visible+Semantic & \begin{tabular}[t]{@{}c@{}}6.14 [4.78, 7.51] \\ 0.46 [0.20, 0.89]\end{tabular} & \begin{tabular}[t]{@{}c@{}}5.55 [4.27, 6.91] \\ 0.37 [0.19, 0.76]\end{tabular} & \begin{tabular}[t]{@{}c@{}}13.24 [11.32, 15.26] \\ 1.51 [0.96, 2.20]\end{tabular} & \begin{tabular}[t]{@{}c@{}}34.47 [31.74, 37.20] \\ 3.08 [2.24, 4.04]\end{tabular} \\
\bottomrule
\end{tabular}
\end{table*}

\section{Paired Comparisons Relative to Direct}
\label{appendix:paired}

Tables~\ref{tab:delta-unc-mmlu} and~\ref{tab:delta-unc-arc} report 95\% confidence intervals for differences relative to Direct, using 10,000 question-level bootstrap resamples. Accuracy comparisons use the full question set, while Flip and RStd comparisons use only questions with at least one parsed rotation under both methods, restricted to four-option questions for RStd. Tables~\ref{tab:mcnemar-mmlu} and~\ref{tab:mcnemar-arc} report the number of questions broken (correct under Direct, incorrect under the other method) and fixed (the reverse), treating unscored questions as incorrect. We use exact two-sided McNemar tests, with Holm correction across the 36 comparisons per benchmark.

\begin{table*}[t]
\centering
\footnotesize
\setlength{\tabcolsep}{4pt}
\caption{Differences relative to the same model's Direct condition on MMLU{,} API models{,} with 95\% paired question-bootstrap CIs (10{,}000 resamples; identical resampled question sets for the condition and Direct; percentile intervals). $\Delta$Accuracy and $\Delta$Flip in percentage points; $\Delta$RStd is the difference in RStd. $\Delta$Accuracy: all questions. $\Delta$Flip and $\Delta$RStd: questions for which the metric is defined under both the condition and Direct (Flip: at least one parsed rotation; RStd: 4-option question with at least one parsed rotation); the bootstrap resamples this shared set.}
\label{tab:delta-unc-mmlu}
\begin{tabular}{@{}lccc@{}}
\toprule
Condition & $\Delta$Accuracy [95\% CI] & $\Delta$Flip [95\% CI] & $\Delta$RStd [95\% CI] \\
\midrule
\multicolumn{4}{@{}l}{\textit{GPT-4.1 mini}} \\
\quad Reasoning Direct & +7.28 [5.26, 9.30] & $-$7.19 [$-$9.65, $-$4.74] & $-$1.41 [$-$2.15, $-$0.50] \\
\quad Two-stage & $-$0.26 [$-$2.37, 1.75] & $-$7.63 [$-$10.09, $-$5.18] & $-$1.57 [$-$2.28, $-$0.67] \\
\quad Reasoning Two-stage & +3.95 [1.75, 6.14] & $-$9.30 [$-$11.75, $-$6.75] & $-$1.24 [$-$2.02, $-$0.37] \\
\quad Visible+LLM & +2.37 [0.70, 4.04] & $-$3.51 [$-$5.61, $-$1.49] & $-$0.73 [$-$1.44, 0.05] \\
\quad Visible+Semantic & +1.58 [$-$0.18, 3.25] & $-$3.33 [$-$5.44, $-$1.23] & $-$0.74 [$-$1.45, 0.01] \\
\multicolumn{4}{@{}l}{\textit{Claude Haiku 4.5}} \\
\quad Reasoning Direct & +2.28 [0.00, 4.47] & $-$1.07 [$-$4.09, 1.87] & +0.27 [$-$0.64, 0.94] \\
\quad Two-stage & $-$1.40 [$-$3.51, 0.70] & $-$5.77 [$-$8.53, $-$3.02] & +1.03 [$-$0.02, 1.59] \\
\quad Reasoning Two-stage & +2.28 [0.18, 4.39] & $-$7.56 [$-$10.22, $-$4.89] & +1.44 [0.37, 2.02] \\
\quad Visible+LLM & $-$1.49 [$-$3.33, 0.26] & $-$2.06 [$-$4.11, 0.09] & +0.48 [$-$0.58, 1.15] \\
\quad Visible+Semantic & +1.75 [$-$0.09, 3.51] & $-$6.50 [$-$8.82, $-$4.10] & +0.39 [$-$0.61, 0.97] \\
\multicolumn{4}{@{}l}{\textit{Qwen2.5-7B}} \\
\quad Reasoning Direct & +0.96 [$-$1.23, 3.25] & +2.54 [$-$0.18, 5.26] & +1.30 [0.13, 2.39] \\
\quad Two-stage & $-$5.79 [$-$8.33, $-$3.25] & +0.53 [$-$2.46, 3.51] & +0.93 [$-$0.21, 1.99] \\
\quad Reasoning Two-stage & $-$3.95 [$-$6.75, $-$1.14] & +0.53 [$-$2.72, 3.86] & +1.06 [$-$0.18, 2.27] \\
\quad Visible+LLM & $-$1.05 [$-$2.63, 0.53] & +0.61 [$-$1.67, 2.81] & +0.22 [$-$0.69, 1.12] \\
\quad Visible+Semantic & $-$15.09 [$-$17.63, $-$12.54] & $-$2.06 [$-$4.78, 0.66] & $-$0.07 [$-$1.21, 1.15] \\
\multicolumn{4}{@{}l}{\textit{Llama-3.1-8B}} \\
\quad Reasoning Direct & +6.49 [3.51, 9.47] & $-$1.49 [$-$4.56, 1.58] & $-$4.75 [$-$5.78, $-$3.13] \\
\quad Two-stage & +0.26 [$-$2.81, 3.33] & $-$5.27 [$-$8.52, $-$2.11] & +3.27 [1.89, 4.57] \\
\quad Reasoning Two-stage & +0.53 [$-$2.63, 3.77] & $-$7.73 [$-$11.06, $-$4.48] & +3.82 [2.42, 5.17] \\
\quad Visible+LLM & +3.95 [1.23, 6.75] & $-$0.96 [$-$3.77, 1.84] & $-$1.31 [$-$2.67, 0.08] \\
\quad Visible+Semantic & $-$0.09 [$-$2.98, 2.81] & $-$3.08 [$-$6.16, 0.00] & $-$1.48 [$-$2.87, $-$0.05] \\
\bottomrule
\end{tabular}
\end{table*}

\begin{table*}[t]
\centering
\footnotesize
\setlength{\tabcolsep}{4pt}
\caption{Differences relative to the same model's Direct condition on ARC-Challenge{,} API models{,} with 95\% paired question-bootstrap CIs (10{,}000 resamples; identical resampled question sets for the condition and Direct; percentile intervals). $\Delta$Accuracy and $\Delta$Flip in percentage points; $\Delta$RStd is the difference in RStd. $\Delta$Accuracy: all questions. $\Delta$Flip and $\Delta$RStd: questions for which the metric is defined under both the condition and Direct (Flip: at least one parsed rotation; RStd: 4-option question with at least one parsed rotation); the bootstrap resamples this shared set.}
\label{tab:delta-unc-arc}
\begin{tabular}{@{}lccc@{}}
\toprule
Condition & $\Delta$Accuracy [95\% CI] & $\Delta$Flip [95\% CI] & $\Delta$RStd [95\% CI] \\
\midrule
\multicolumn{4}{@{}l}{\textit{GPT-4.1 mini}} \\
\quad Reasoning Direct & +0.34 [$-$0.68, 1.37] & $-$2.30 [$-$3.75, $-$0.94] & $-$0.06 [$-$0.45, 0.26] \\
\quad Two-stage & $-$1.19 [$-$2.39, 0.00] & $-$0.17 [$-$1.71, 1.45] & +0.21 [$-$0.29, 0.56] \\
\quad Reasoning Two-stage & $-$1.54 [$-$2.82, $-$0.26] & $-$0.68 [$-$2.22, 0.85] & +0.29 [$-$0.24, 0.61] \\
\quad Visible+LLM & +0.26 [$-$0.43, 0.94] & $-$0.17 [$-$1.37, 0.94] & +0.25 [$-$0.22, 0.55] \\
\quad Visible+Semantic & +0.09 [$-$0.68, 0.85] & +0.09 [$-$1.11, 1.28] & +0.28 [$-$0.20, 0.59] \\
\multicolumn{4}{@{}l}{\textit{Claude Haiku 4.5}} \\
\quad Reasoning Direct & $-$0.26 [$-$1.62, 1.11] & +2.65 [0.77, 4.61] & $-$0.05 [$-$0.61, 0.57] \\
\quad Two-stage & $-$2.05 [$-$3.58, $-$0.51] & +0.94 [$-$0.94, 2.73] & +0.07 [$-$0.61, 0.73] \\
\quad Reasoning Two-stage & $-$2.39 [$-$4.01, $-$0.85] & +1.11 [$-$0.77, 2.99] & +0.27 [$-$0.44, 0.93] \\
\quad Visible+LLM & $-$0.26 [$-$1.19, 0.68] & $-$2.57 [$-$3.77, $-$1.46] & $-$0.28 [$-$0.70, 0.16] \\
\quad Visible+Semantic & +0.00 [$-$1.02, 1.02] & $-$2.82 [$-$4.18, $-$1.54] & $-$0.30 [$-$0.73, 0.17] \\
\multicolumn{4}{@{}l}{\textit{Qwen2.5-7B}} \\
\quad Reasoning Direct & +1.54 [0.09, 3.07] & +0.68 [$-$1.45, 2.73] & +1.23 [0.41, 1.93] \\
\quad Two-stage & $-$5.38 [$-$7.34, $-$3.33] & +5.80 [3.24, 8.36] & +1.75 [0.80, 2.60] \\
\quad Reasoning Two-stage & $-$8.70 [$-$11.09, $-$6.40] & +12.46 [9.64, 15.27] & +4.50 [3.36, 5.50] \\
\quad Visible+LLM & $-$0.17 [$-$1.19, 0.85] & +0.43 [$-$1.02, 1.88] & +0.41 [$-$0.17, 0.94] \\
\quad Visible+Semantic & $-$10.58 [$-$12.63, $-$8.53] & $-$0.61 [$-$2.35, 1.13] & +0.65 [$-$0.05, 1.28] \\
\multicolumn{4}{@{}l}{\textit{Llama-3.1-8B}} \\
\quad Reasoning Direct & +6.40 [3.92, 8.87] & +1.28 [$-$1.62, 4.27] & $-$0.89 [$-$1.95, 0.29] \\
\quad Two-stage & $-$7.25 [$-$9.90, $-$4.52] & +6.31 [3.07, 9.47] & +5.29 [4.03, 6.43] \\
\quad Reasoning Two-stage & $-$3.58 [$-$6.40, $-$0.68] & +6.23 [2.82, 9.47] & +6.59 [5.31, 7.76] \\
\quad Visible+LLM & +0.68 [$-$1.62, 2.99] & +2.82 [$-$0.09, 5.80] & +1.09 [0.03, 2.14] \\
\quad Visible+Semantic & $-$1.28 [$-$3.75, 1.19] & +4.52 [1.37, 7.76] & +0.82 [$-$0.27, 1.94] \\
\bottomrule
\end{tabular}
\end{table*}

\begin{table*}[t]
\centering
\footnotesize
\setlength{\tabcolsep}{4pt}
\caption{Paired per-question comparison of each condition against the same model's Direct condition on MMLU{,} API models. Broken: correct under Direct{,} incorrect under the condition; Fixed: incorrect under Direct{,} correct under the condition (unscored counts as incorrect). $n$: paired questions (all conditions cover the identical question set). $\Delta$Acc $=$ (Fixed $-$ Broken)$/n${,} in percentage points{,} with 95\% paired question-bootstrap CI (10{,}000 resamples). $p$: exact two-sided McNemar test (binomial test on discordant pairs). $p_{\mathrm{Holm}}$: Holm-adjusted over the family of all 36 condition-vs-Direct tests on MMLU (4 models $\times$ 9 conditions).}
\label{tab:mcnemar-mmlu}
\begin{tabular}{@{}lrrrrcrr@{}}
\toprule
Condition & Broken & Fixed & $n$ & $\Delta$Acc & 95\% CI & $p$ & $p_{\mathrm{Holm}}$ \\
\midrule
\multicolumn{8}{@{}l}{\textit{GPT-4.1 mini}} \\
\quad Reasoning Direct & 32 & 115 & 1140 & +7.28 & [5.26, 9.30] & $3.46\times10^{-12}$ & $1.04\times10^{-10}$ \\
\quad Cyclic & 25 & 28 & 1140 & +0.26 & [$-$0.96, 1.49] & 0.784 & 1.000 \\
\quad Reasoning Cyclic & 33 & 114 & 1140 & +7.11 & [5.00, 9.21] & $1.22\times10^{-11}$ & $3.54\times10^{-10}$ \\
\quad Two-stage & 73 & 70 & 1140 & $-$0.26 & [$-$2.37, 1.75] & 0.867 & 1.000 \\
\quad Reasoning Two-stage & 59 & 104 & 1140 & +3.95 & [1.75, 6.14] & $5.28\times10^{-4}$ & 0.013 \\
\quad IHS & 96 & 104 & 1140 & +0.70 & [$-$1.75, 3.16] & 0.621 & 1.000 \\
\quad Hidden+Semantic & 454 & 53 & 1140 & $-$35.18 & [$-$38.42, $-$31.93] & $1.75\times10^{-80}$ & $6.13\times10^{-79}$ \\
\quad Visible+LLM & 33 & 60 & 1140 & +2.37 & [0.70, 4.04] & 0.007 & 0.147 \\
\quad Visible+Semantic & 41 & 59 & 1140 & +1.58 & [$-$0.18, 3.25] & 0.089 & 1.000 \\
\multicolumn{8}{@{}l}{\textit{Claude Haiku 4.5}} \\
\quad Reasoning Direct & 72 & 98 & 1140 & +2.28 & [0.00, 4.47] & 0.055 & 0.933 \\
\quad Cyclic & 34 & 45 & 1140 & +0.96 & [$-$0.61, 2.54] & 0.260 & 1.000 \\
\quad Reasoning Cyclic & 38 & 104 & 1140 & +5.79 & [3.77, 7.81] & $2.78\times10^{-8}$ & $7.79\times10^{-7}$ \\
\quad Two-stage & 84 & 68 & 1140 & $-$1.40 & [$-$3.51, 0.70] & 0.224 & 1.000 \\
\quad Reasoning Two-stage & 65 & 91 & 1140 & +2.28 & [0.18, 4.39] & 0.045 & 0.810 \\
\quad IHS & 84 & 100 & 1140 & +1.40 & [$-$0.96, 3.68] & 0.269 & 1.000 \\
\quad Hidden+Semantic & 488 & 35 & 1140 & $-$39.74 & [$-$42.89, $-$36.58] & $3.33\times10^{-103}$ & $1.20\times10^{-101}$ \\
\quad Visible+LLM & 63 & 46 & 1140 & $-$1.49 & [$-$3.33, 0.26] & 0.125 & 1.000 \\
\quad Visible+Semantic & 46 & 66 & 1140 & +1.75 & [$-$0.09, 3.51] & 0.072 & 1.000 \\
\multicolumn{8}{@{}l}{\textit{Qwen2.5-7B}} \\
\quad Reasoning Direct & 78 & 89 & 1140 & +0.96 & [$-$1.23, 3.25] & 0.439 & 1.000 \\
\quad Cyclic & 43 & 53 & 1140 & +0.88 & [$-$0.79, 2.63] & 0.358 & 1.000 \\
\quad Reasoning Cyclic & 69 & 116 & 1140 & +4.12 & [1.75, 6.49] & $6.78\times10^{-4}$ & 0.016 \\
\quad Two-stage & 146 & 80 & 1140 & $-$5.79 & [$-$8.33, $-$3.25] & $1.34\times10^{-5}$ & $3.63\times10^{-4}$ \\
\quad Reasoning Two-stage & 155 & 110 & 1140 & $-$3.95 & [$-$6.75, $-$1.14] & 0.007 & 0.147 \\
\quad IHS & 194 & 117 & 1140 & $-$6.75 & [$-$9.74, $-$3.77] & $1.49\times10^{-5}$ & $3.87\times10^{-4}$ \\
\quad Hidden+Semantic & 455 & 72 & 1140 & $-$33.60 & [$-$37.02, $-$30.26] & $5.10\times10^{-69}$ & $1.73\times10^{-67}$ \\
\quad Visible+LLM & 49 & 37 & 1140 & $-$1.05 & [$-$2.63, 0.53] & 0.235 & 1.000 \\
\quad Visible+Semantic & 205 & 33 & 1140 & $-$15.09 & [$-$17.63, $-$12.54] & $1.62\times10^{-31}$ & $5.34\times10^{-30}$ \\
\multicolumn{8}{@{}l}{\textit{Llama-3.1-8B}} \\
\quad Reasoning Direct & 119 & 193 & 1140 & +6.49 & [3.51, 9.47] & $3.31\times10^{-5}$ & $8.28\times10^{-4}$ \\
\quad Cyclic & 66 & 98 & 1140 & +2.81 & [0.61, 5.00] & 0.015 & 0.289 \\
\quad Reasoning Cyclic & 92 & 224 & 1140 & +11.58 & [8.68, 14.56] & $7.43\times10^{-14}$ & $2.30\times10^{-12}$ \\
\quad Two-stage & 155 & 158 & 1140 & +0.26 & [$-$2.81, 3.33] & 0.910 & 1.000 \\
\quad Reasoning Two-stage & 166 & 172 & 1140 & +0.53 & [$-$2.63, 3.77] & 0.786 & 1.000 \\
\quad IHS & 223 & 186 & 1140 & $-$3.25 & [$-$6.67, 0.18] & 0.075 & 1.000 \\
\quad Hidden+Semantic & 367 & 119 & 1140 & $-$21.75 & [$-$25.26, $-$18.07] & $1.89\times10^{-30}$ & $6.03\times10^{-29}$ \\
\quad Visible+LLM & 111 & 156 & 1140 & +3.95 & [1.23, 6.75] & 0.007 & 0.147 \\
\quad Visible+Semantic & 150 & 149 & 1140 & $-$0.09 & [$-$2.98, 2.81] & 1.000 & 1.000 \\
\bottomrule
\end{tabular}
\end{table*}

\begin{table*}[t]
\centering
\footnotesize
\setlength{\tabcolsep}{4pt}
\caption{Paired per-question comparison of each condition against the same model's Direct condition on ARC-Challenge{,} API models. Broken: correct under Direct{,} incorrect under the condition; Fixed: incorrect under Direct{,} correct under the condition (unscored counts as incorrect). $n$: paired questions (all conditions cover the identical question set). $\Delta$Acc $=$ (Fixed $-$ Broken)$/n${,} in percentage points{,} with 95\% paired question-bootstrap CI (10{,}000 resamples). $p$: exact two-sided McNemar test (binomial test on discordant pairs). $p_{\mathrm{Holm}}$: Holm-adjusted over the family of all 36 condition-vs-Direct tests on ARC-Challenge (4 models $\times$ 9 conditions).}
\label{tab:mcnemar-arc}
\begin{tabular}{@{}lrrrrcrr@{}}
\toprule
Condition & Broken & Fixed & $n$ & $\Delta$Acc & 95\% CI & $p$ & $p_{\mathrm{Holm}}$ \\
\midrule
\multicolumn{8}{@{}l}{\textit{GPT-4.1 mini}} \\
\quad Reasoning Direct & 16 & 20 & 1172 & +0.34 & [$-$0.68, 1.37] & 0.618 & 1.000 \\
\quad Cyclic & 7 & 11 & 1172 & +0.34 & [$-$0.34, 1.02] & 0.481 & 1.000 \\
\quad Reasoning Cyclic & 7 & 24 & 1172 & +1.45 & [0.51, 2.39] & 0.003 & 0.073 \\
\quad Two-stage & 33 & 19 & 1172 & $-$1.19 & [$-$2.39, 0.00] & 0.070 & 0.845 \\
\quad Reasoning Two-stage & 40 & 22 & 1172 & $-$1.54 & [$-$2.82, $-$0.26] & 0.030 & 0.450 \\
\quad IHS & 49 & 26 & 1172 & $-$1.96 & [$-$3.41, $-$0.60] & 0.011 & 0.190 \\
\quad Hidden+Semantic & 510 & 13 & 1172 & $-$42.41 & [$-$45.31, $-$39.42] & $2.26\times10^{-132}$ & $7.92\times10^{-131}$ \\
\quad Visible+LLM & 8 & 11 & 1172 & +0.26 & [$-$0.43, 0.94] & 0.648 & 1.000 \\
\quad Visible+Semantic & 11 & 12 & 1172 & +0.09 & [$-$0.68, 0.85] & 1.000 & 1.000 \\
\multicolumn{8}{@{}l}{\textit{Claude Haiku 4.5}} \\
\quad Reasoning Direct & 36 & 33 & 1172 & $-$0.26 & [$-$1.62, 1.11] & 0.810 & 1.000 \\
\quad Cyclic & 9 & 21 & 1172 & +1.02 & [0.09, 1.96] & 0.043 & 0.599 \\
\quad Reasoning Cyclic & 16 & 37 & 1172 & +1.79 & [0.60, 3.07] & 0.005 & 0.110 \\
\quad Two-stage & 54 & 30 & 1172 & $-$2.05 & [$-$3.58, $-$0.51] & 0.012 & 0.198 \\
\quad Reasoning Two-stage & 58 & 30 & 1172 & $-$2.39 & [$-$4.01, $-$0.85] & 0.004 & 0.079 \\
\quad IHS & 28 & 34 & 1172 & +0.51 & [$-$0.77, 1.79] & 0.526 & 1.000 \\
\quad Hidden+Semantic & 526 & 14 & 1172 & $-$43.69 & [$-$46.59, $-$40.70] & $9.90\times10^{-136}$ & $3.57\times10^{-134}$ \\
\quad Visible+LLM & 18 & 15 & 1172 & $-$0.26 & [$-$1.19, 0.68] & 0.728 & 1.000 \\
\quad Visible+Semantic & 19 & 19 & 1172 & +0.00 & [$-$1.02, 1.02] & 1.000 & 1.000 \\
\multicolumn{8}{@{}l}{\textit{Qwen2.5-7B}} \\
\quad Reasoning Direct & 31 & 49 & 1172 & +1.54 & [0.09, 3.07] & 0.057 & 0.737 \\
\quad Cyclic & 9 & 26 & 1172 & +1.45 & [0.51, 2.47] & 0.006 & 0.114 \\
\quad Reasoning Cyclic & 20 & 62 & 1172 & +3.58 & [2.13, 5.12] & $3.71\times10^{-6}$ & $9.66\times10^{-5}$ \\
\quad Two-stage & 106 & 43 & 1172 & $-$5.38 & [$-$7.34, $-$3.33] & $2.54\times10^{-7}$ & $7.37\times10^{-6}$ \\
\quad Reasoning Two-stage & 153 & 51 & 1172 & $-$8.70 & [$-$11.09, $-$6.40] & $4.93\times10^{-13}$ & $1.48\times10^{-11}$ \\
\quad IHS & 124 & 62 & 1172 & $-$5.29 & [$-$7.59, $-$3.07] & $6.41\times10^{-6}$ & $1.60\times10^{-4}$ \\
\quad Hidden+Semantic & 518 & 34 & 1172 & $-$41.30 & [$-$44.37, $-$38.23] & $2.93\times10^{-112}$ & $9.97\times10^{-111}$ \\
\quad Visible+LLM & 20 & 18 & 1172 & $-$0.17 & [$-$1.19, 0.85] & 0.871 & 1.000 \\
\quad Visible+Semantic & 142 & 18 & 1172 & $-$10.58 & [$-$12.63, $-$8.53] & $4.27\times10^{-25}$ & $1.37\times10^{-23}$ \\
\multicolumn{8}{@{}l}{\textit{Llama-3.1-8B}} \\
\quad Reasoning Direct & 73 & 148 & 1172 & +6.40 & [3.92, 8.87] & $5.05\times10^{-7}$ & $1.36\times10^{-5}$ \\
\quad Cyclic & 28 & 68 & 1172 & +3.41 & [1.79, 5.12] & $5.46\times10^{-5}$ & 0.001 \\
\quad Reasoning Cyclic & 37 & 168 & 1172 & +11.18 & [8.96, 13.48] & $3.88\times10^{-21}$ & $1.20\times10^{-19}$ \\
\quad Two-stage & 178 & 93 & 1172 & $-$7.25 & [$-$9.90, $-$4.52] & $2.70\times10^{-7}$ & $7.56\times10^{-6}$ \\
\quad Reasoning Two-stage & 171 & 129 & 1172 & $-$3.58 & [$-$6.40, $-$0.68] & 0.018 & 0.284 \\
\quad IHS & 184 & 117 & 1172 & $-$5.72 & [$-$8.53, $-$2.82] & $1.34\times10^{-4}$ & 0.003 \\
\quad Hidden+Semantic & 488 & 85 & 1172 & $-$34.39 & [$-$37.88, $-$30.89] & $1.09\times10^{-69}$ & $3.60\times10^{-68}$ \\
\quad Visible+LLM & 89 & 97 & 1172 & +0.68 & [$-$1.62, 2.99] & 0.608 & 1.000 \\
\quad Visible+Semantic & 117 & 102 & 1172 & $-$1.28 & [$-$3.75, 1.19] & 0.344 & 1.000 \\
\bottomrule
\end{tabular}
\end{table*}

\section{\texorpdfstring{$2\times2$}{2x2} Diagnostic Contrasts}
\label{appendix:grid}

Tables~\ref{tab:2x2-contrasts-mmlu} and~\ref{tab:2x2-contrasts-arc} report the four paired contrasts between the diagnostic cells in Tables~\ref{tab:2x2-mmlu} and~\ref{tab:2x2-arc}: Hidden+LLM $-$ Visible+LLM, Hidden+Semantic $-$ Visible+Semantic, Hidden+LLM $-$ Hidden+Semantic, and Visible+LLM $-$ Visible+Semantic. $\Delta$Accuracy uses all questions. $\Delta$Flip and $\Delta$RStd use the questions for which the metric is defined under both cells, with the same paired bootstrap as Appendix~\ref{appendix:paired}. Hidden+Semantic matches the Two-stage stage-1 text, which is generated without options, to each rotation of the options. Its per-rotation choices are identical for every question, so its Flip and RStd are 0.

\begin{table*}[t]
\centering
\footnotesize
\setlength{\tabcolsep}{4pt}
\caption{Paired contrasts between $2\times2$ diagnostic cells on MMLU{,} API models (first minus second){,} with 95\% paired question-bootstrap CIs (10{,}000 resamples{,} percentile). $\Delta$Accuracy and $\Delta$Flip in percentage points; $\Delta$RStd is the difference in RStd. $\Delta$Accuracy: all questions. $\Delta$Flip and $\Delta$RStd: questions for which the metric is defined under both cells (Flip: at least one parsed rotation; RStd: 4-option question with at least one parsed rotation); the bootstrap resamples this shared set.}
\label{tab:2x2-contrasts-mmlu}
\begin{tabular}{@{}lccc@{}}
\toprule
Contrast & $\Delta$Accuracy [95\% CI] & $\Delta$Flip [95\% CI] & $\Delta$RStd [95\% CI] \\
\midrule
\multicolumn{4}{@{}l}{\textit{GPT-4.1 mini}} \\
\quad Hidden+LLM $-$ Visible+LLM & $-$2.63 [$-$4.56, $-$0.70] & $-$4.12 [$-$6.49, $-$1.75] & $-$0.84 [$-$1.46, $-$0.02] \\
\quad Hidden+Semantic $-$ Visible+Semantic & $-$36.75 [$-$39.91, $-$33.60] & $-$15.84 [$-$18.17, $-$13.60] & $-$1.21 [$-$1.90, $-$0.70] \\
\quad Hidden+LLM $-$ Hidden+Semantic & +34.91 [31.84, 37.98] & +11.47 [9.54, 13.50] & +0.38 [0.17, 1.04] \\
\quad Visible+LLM $-$ Visible+Semantic & +0.79 [0.18, 1.40] & $-$0.18 [$-$0.70, 0.26] & +0.01 [$-$0.18, 0.19] \\
\multicolumn{4}{@{}l}{\textit{Claude Haiku 4.5}} \\
\quad Hidden+LLM $-$ Visible+LLM & +0.09 [$-$2.19, 2.37] & $-$3.91 [$-$6.57, $-$1.15] & +0.55 [$-$0.44, 1.37] \\
\quad Hidden+Semantic $-$ Visible+Semantic & $-$41.49 [$-$44.56, $-$38.42] & $-$14.52 [$-$16.83, $-$12.32] & $-$0.63 [$-$1.35, $-$0.27] \\
\quad Hidden+LLM $-$ Hidden+Semantic & +38.33 [35.35, 41.40] & +13.85 [11.65, 16.15] & +1.20 [0.71, 1.88] \\
\quad Visible+LLM $-$ Visible+Semantic & $-$3.25 [$-$4.74, $-$1.75] & +4.28 [2.85, 5.70] & $-$0.01 [$-$0.85, 0.92] \\
\multicolumn{4}{@{}l}{\textit{Qwen2.5-7B}} \\
\quad Hidden+LLM $-$ Visible+LLM & $-$4.74 [$-$7.37, $-$2.02] & $-$0.09 [$-$3.07, 2.89] & +0.70 [$-$0.47, 1.84] \\
\quad Hidden+Semantic $-$ Visible+Semantic & $-$18.51 [$-$22.11, $-$15.00] & $-$28.24 [$-$31.21, $-$25.38] & $-$2.19 [$-$3.28, $-$1.37] \\
\quad Hidden+LLM $-$ Hidden+Semantic & +27.81 [24.56, 30.96] & +28.78 [26.00, 31.65] & +3.53 [2.70, 4.45] \\
\quad Visible+LLM $-$ Visible+Semantic & +14.04 [11.93, 16.14] & +2.35 [0.47, 4.32] & +0.15 [$-$0.74, 0.96] \\
\multicolumn{4}{@{}l}{\textit{Llama-3.1-8B}} \\
\quad Hidden+LLM $-$ Visible+LLM & $-$3.68 [$-$6.75, $-$0.61] & $-$4.31 [$-$7.56, $-$1.14] & +4.48 [3.09, 5.84] \\
\quad Hidden+Semantic $-$ Visible+Semantic & $-$21.67 [$-$25.09, $-$18.25] & $-$49.73 [$-$52.93, $-$46.53] & $-$4.18 [$-$5.38, $-$3.14] \\
\quad Hidden+LLM $-$ Hidden+Semantic & +22.02 [18.86, 25.18] & +44.30 [41.21, 47.50] & +8.26 [7.17, 9.40] \\
\quad Visible+LLM $-$ Visible+Semantic & +4.04 [2.37, 5.70] & +2.02 [0.09, 3.96] & +0.18 [$-$0.60, 0.94] \\
\bottomrule
\end{tabular}
\end{table*}

\begin{table*}[t]
\centering
\footnotesize
\setlength{\tabcolsep}{4pt}
\caption{Paired contrasts between $2\times2$ diagnostic cells on ARC-Challenge{,} API models (first minus second){,} with 95\% paired question-bootstrap CIs (10{,}000 resamples{,} percentile). $\Delta$Accuracy and $\Delta$Flip in percentage points; $\Delta$RStd is the difference in RStd. $\Delta$Accuracy: all questions. $\Delta$Flip and $\Delta$RStd: questions for which the metric is defined under both cells (Flip: at least one parsed rotation; RStd: 4-option question with at least one parsed rotation); the bootstrap resamples this shared set.}
\label{tab:2x2-contrasts-arc}
\begin{tabular}{@{}lccc@{}}
\toprule
Contrast & $\Delta$Accuracy [95\% CI] & $\Delta$Flip [95\% CI] & $\Delta$RStd [95\% CI] \\
\midrule
\multicolumn{4}{@{}l}{\textit{GPT-4.1 mini}} \\
\quad Hidden+LLM $-$ Visible+LLM & $-$1.45 [$-$2.73, $-$0.17] & +0.00 [$-$1.54, 1.45] & $-$0.05 [$-$0.52, 0.44] \\
\quad Hidden+Semantic $-$ Visible+Semantic & $-$42.49 [$-$45.48, $-$39.51] & $-$5.85 [$-$7.29, $-$4.41] & $-$0.32 [$-$0.78, $-$0.14] \\
\quad Hidden+LLM $-$ Hidden+Semantic & +41.21 [38.31, 44.11] & +4.80 [3.55, 6.14] & +0.29 [0.13, 0.70] \\
\quad Visible+LLM $-$ Visible+Semantic & +0.17 [$-$0.17, 0.51] & $-$0.26 [$-$0.68, 0.09] & $-$0.02 [$-$0.13, 0.09] \\
\multicolumn{4}{@{}l}{\textit{Claude Haiku 4.5}} \\
\quad Hidden+LLM $-$ Visible+LLM & $-$1.79 [$-$3.24, $-$0.34] & +3.60 [1.80, 5.40] & +0.35 [$-$0.24, 0.91] \\
\quad Hidden+Semantic $-$ Visible+Semantic & $-$43.69 [$-$46.67, $-$40.70] & $-$4.78 [$-$6.07, $-$3.48] & $-$0.22 [$-$0.64, $-$0.11] \\
\quad Hidden+LLM $-$ Hidden+Semantic & +41.64 [38.74, 44.54] & +7.76 [6.17, 9.45] & +0.52 [0.19, 1.07] \\
\quad Visible+LLM $-$ Visible+Semantic & $-$0.26 [$-$0.85, 0.34] & +0.17 [$-$0.43, 0.77] & +0.02 [$-$0.18, 0.18] \\
\multicolumn{4}{@{}l}{\textit{Qwen2.5-7B}} \\
\quad Hidden+LLM $-$ Visible+LLM & $-$5.20 [$-$7.25, $-$3.16] & +5.38 [2.82, 7.94] & +1.34 [0.42, 2.20] \\
\quad Hidden+Semantic $-$ Visible+Semantic & $-$30.72 [$-$34.04, $-$27.39] & $-$12.30 [$-$14.38, $-$10.22] & $-$1.23 [$-$1.93, $-$0.71] \\
\quad Hidden+LLM $-$ Hidden+Semantic & +35.92 [32.94, 38.82] & +17.23 [15.00, 19.47] & +2.16 [1.53, 2.90] \\
\quad Visible+LLM $-$ Visible+Semantic & +10.41 [8.62, 12.29] & +1.05 [0.00, 2.09] & $-$0.25 [$-$0.70, 0.19] \\
\multicolumn{4}{@{}l}{\textit{Llama-3.1-8B}} \\
\quad Hidden+LLM $-$ Visible+LLM & $-$7.94 [$-$10.58, $-$5.20] & +3.50 [0.34, 6.74] & +4.19 [2.93, 5.39] \\
\quad Hidden+Semantic $-$ Visible+Semantic & $-$33.11 [$-$36.43, $-$29.69] & $-$33.13 [$-$36.13, $-$30.24] & $-$2.91 [$-$3.93, $-$2.06] \\
\quad Hidden+LLM $-$ Hidden+Semantic & +27.13 [24.06, 30.38] & +32.53 [29.64, 35.43] & +6.67 [5.72, 7.66] \\
\quad Visible+LLM $-$ Visible+Semantic & +1.96 [0.77, 3.24] & $-$1.71 [$-$3.24, $-$0.17] & +0.27 [$-$0.24, 0.76] \\
\bottomrule
\end{tabular}
\end{table*}

\section{Additional Robustness Analyses}
\label{appendix:results}

Holm-corrected exact McNemar tests against Direct are reported in Appendix~\ref{appendix:paired} (Tables~\ref{tab:mcnemar-mmlu} and~\ref{tab:mcnemar-arc}).

\paragraph{Repeated runs.} Table~\ref{tab:stochasticity} reports repeated-run stability for Direct, Reasoning Direct, Two-stage and Reasoning Two-stage on a fixed 100-question subset per benchmark, for the four API models. Repetition~0 is the main-run observation; repetitions 1--3 are fresh requests. The table reports the number of questions with an identical parsed choice in all four repetitions, together with scored and correct counts per repetition. The analysis uses counts only and no bootstrap. The subset is selected deterministically from the evaluation set of each benchmark: questions are ranked by a hash of the seed (42), benchmark and question ID, and the first 100 are taken.

\paragraph{Local verification.} Table~\ref{tab:local} reports E2E accuracy for local runs of \texttt{RedHatAI/\allowbreak Qwen2.5-7B-Instruct-\allowbreak FP8-dynamic} and \texttt{RedHatAI/\allowbreak Meta-Llama-3.1-8B-\allowbreak Instruct-FP8-dynamic}. Both use batched Hugging Face inference (batch size 32, fast FP8 quantize--dequantize), greedy decoding and a 1024-token cap. 

\begin{table*}[t]
\centering
\footnotesize
\setlength{\tabcolsep}{3pt}
\caption{Repeated-run subset (100 questions per benchmark; 4 repetitions: repetition 0 is the main run, repetitions 1--3 are fresh runs). Ident.: questions with identical parsed choice in all four repetitions. Scored and Correct: counts per repetition 0/1/2/3.}
\label{tab:stochasticity}
\begin{tabular}{@{}llcccccc@{}}
\toprule
 & & \multicolumn{3}{c}{MMLU} & \multicolumn{3}{c}{ARC-Challenge} \\
\cmidrule(lr){3-5}\cmidrule(lr){6-8}
Model & Method & Ident. & Scored & Correct & Ident. & Scored & Correct \\
\midrule
GPT-4.1 mini & Direct & 100/100 & 100/100/100/100 & 81/81/81/81 & 100/100 & 100/100/100/100 & 98/98/98/98 \\
 & Reasoning Direct & 96/100 & 100/100/100/99 & 92/89/90/90 & 98/100 & 99/99/99/99 & 97/96/96/96 \\
 & Two-stage & 95/100 & 99/100/100/100 & 82/83/83/82 & 97/100 & 100/100/100/100 & 94/92/93/93 \\
 & Reasoning Two-stage & 93/100 & 100/100/100/100 & 90/86/88/87 & 96/100 & 100/100/100/100 & 94/95/93/95 \\
Claude Haiku 4.5 & Direct & 97/100 & 97/96/97/97 & 84/85/83/84 & 100/100 & 100/100/100/100 & 95/95/95/95 \\
 & Reasoning Direct & 88/100 & 99/99/98/99 & 88/88/89/87 & 95/100 & 100/100/100/100 & 93/94/93/93 \\
 & Two-stage & 95/100 & 98/98/98/98 & 82/82/81/83 & 98/100 & 100/100/100/100 & 91/92/92/91 \\
 & Reasoning Two-stage & 88/100 & 97/96/96/99 & 88/84/84/86 & 94/100 & 100/100/100/100 & 95/94/94/95 \\
Qwen2.5-7B & Direct & 99/100 & 100/100/100/100 & 73/73/73/73 & 100/100 & 100/100/100/100 & 87/87/87/87 \\
 & Reasoning Direct & 79/100 & 100/100/100/100 & 78/80/75/74 & 92/100 & 100/100/99/100 & 94/91/91/92 \\
 & Two-stage & 92/100 & 100/100/100/100 & 65/65/67/65 & 98/100 & 100/100/100/100 & 84/84/83/83 \\
 & Reasoning Two-stage & 78/100 & 100/100/100/100 & 71/72/71/73 & 80/100 & 100/100/100/100 & 83/82/80/81 \\
Llama-3.1-8B & Direct & 90/100 & 100/100/100/100 & 53/53/52/54 & 99/100 & 100/100/100/100 & 77/77/77/78 \\
 & Reasoning Direct & 73/100 & 100/100/100/100 & 70/71/65/69 & 93/100 & 100/100/100/100 & 86/82/85/84 \\
 & Two-stage & 77/100 & 98/98/98/99 & 62/58/53/59 & 85/100 & 100/100/100/100 & 66/65/70/68 \\
 & Reasoning Two-stage & 69/100 & 100/100/100/100 & 60/59/57/58 & 80/100 & 100/100/100/100 & 75/72/75/77 \\
\bottomrule
\end{tabular}
\end{table*}

\begin{table*}[t]
\centering
\small
\caption{Local verification runs (RedHatAI FP8 checkpoints, batched HF inference, greedy decoding): end-to-end accuracy (\%). Parenthesized: scored/total, shown when it differs from the full benchmark size. -- : no canonical local result exists in the data bundle.}
\label{tab:local}
\begin{tabular}{lrrrr}
\toprule
 & \multicolumn{2}{c}{MMLU} & \multicolumn{2}{c}{ARC-Challenge} \\
\cmidrule(lr){2-3}\cmidrule(lr){4-5}
Method & Qwen2.5-7B FP8 & Llama-3.1-8B FP8 & Qwen2.5-7B FP8 & Llama-3.1-8B FP8 \\
\midrule
Direct & -- & -- & -- & -- \\
Reasoning Direct & -- & -- & -- & -- \\
Cyclic & 75.53 & 59.21 & 90.70 & 68.26\,{\scriptsize(1171/1172)} \\
Reasoning Cyclic & 79.39 & 73.68 & 92.15 & 89.51 \\
Two-stage & 63.68\,{\scriptsize(1138/1140)} & 52.72\,{\scriptsize(1107/1140)} & 78.41\,{\scriptsize(1170/1172)} & 61.18\,{\scriptsize(1146/1172)} \\
Reasoning Two-stage & 62.46\,{\scriptsize(1137/1140)} & 61.40\,{\scriptsize(1131/1140)} & 70.56\,{\scriptsize(1168/1172)} & 74.32\,{\scriptsize(1165/1172)} \\
IHS & 62.81 & 51.84\,{\scriptsize(1137/1140)} & 77.99 & 68.94 \\
Hidden+Semantic & 38.07\,{\scriptsize(903/1140)} & 31.75\,{\scriptsize(793/1140)} & 47.18\,{\scriptsize(1018/1172)} & 34.64\,{\scriptsize(903/1172)} \\
Visible+LLM & -- & -- & -- & -- \\
Visible+Semantic & 67.63\,{\scriptsize(1076/1140)} & 49.74\,{\scriptsize(1035/1140)} & 84.73\,{\scriptsize(1137/1172)} & 65.27\,{\scriptsize(1123/1172)} \\
\bottomrule
\end{tabular}
\par\vspace{2pt}
{\footnotesize Accuracy counted from the per-question result files (no evaluation reports exist for local runs). Visible+Semantic uses the corrected cascade re-parse files. Local Direct, Reasoning Direct, Visible+LLM, flip/RStd and stochasticity were not completed and are absent.}
\end{table*}

\section{Prompt Templates and Implementation Details}
\label{appendix:implementation}

\subsection{Prompt templates}
\label{appendix:prompts}

All requests are single-turn user messages without a system prompt. \texttt{\{question\}} is the question text and \texttt{\{options\}} is one line per option of the form \texttt{A. <option text>}. Three prompt bundles are used. \texttt{v1} serves Direct, cyclic permutation, Two-stage, Visible+LLM and Visible+Semantic; \texttt{v1\_reasoning} serves Reasoning Direct, Reasoning Cyclic and Reasoning Two-stage; \texttt{v1\_ihs} serves IHS. The templates below are copied verbatim from the prompt snapshots stored in the canonical run directories.

\paragraph{Direct and cyclic permutation (\texttt{v1/direct\_mcq}).}
\begin{PromptBox}
Answer the following multiple-choice question.

Question: {question}

Options:
{options}

Respond with only the letter.
\end{PromptBox}

\paragraph{Reasoning Direct and Reasoning Cyclic (\texttt{v1\_reasoning/direct\_mcq}).}
\begin{PromptBox}
Answer the following multiple-choice question.

Question: {question}

Options:
{options}

First, briefly reason step by step about which option 
is correct. Then, on a new line, state your final 
answer as: 

Final answer: X
\end{PromptBox}

\paragraph{Two-stage, Stage 1 (\texttt{v1/free\_text}; options hidden).}
\begin{PromptBox}
Answer the following question based on your 
knowledge.

Question: {question}

Respond with a short direct answer only.
\end{PromptBox}

\paragraph{Reasoning Two-stage, Stage 1 (\texttt{v1\_reasoning/free\_text}; options hidden).}
\begin{PromptBox}
Answer the following question based on your 
knowledge.

Question: {question}

First, briefly reason step by step about the 
question. Then, on a new line, state your 
answer as: Answer: <your answer>
\end{PromptBox}

\paragraph{LLM matcher (\texttt{option\_matching}; identical in \texttt{v1} and \texttt{v1\_reasoning}).} Used as Stage 2 of Two-stage, Reasoning Two-stage and Visible+LLM. \texttt{\{free\_text\}} is the full Stage 1 response: the hidden-option answer for Two-stage and Reasoning Two-stage, and the text-extraction answer for Visible+LLM.
\begin{PromptBox}
You are given a question, a reference answer, 
and the options.

Question: {question}

Reference answer: {free_text}

Options:
{options}

Select the option that best matches the 
reference answer in the context of the 
question. If the reference answer is 
imperfect or incomplete, choose the closest 
option.

Respond with only the letter.
\end{PromptBox}

\paragraph{Text extraction (\texttt{v1/text\_extraction}; options visible).} Stage 1 of Visible+Semantic and Visible+LLM.
\begin{PromptBox}
Answer the following question. Read all 
options carefully.

Question: {question}

Options:
{options}

Respond with the correct answer text only.
Do not write the option letter. Do not 
explain.
\end{PromptBox}

\paragraph{IHS (\texttt{v1\_ihs/independent\_hypothesis}; one request per option).}
\begin{PromptBox}
Question: {question}

Hypothesis: The correct answer is {option_text}.

Task: Please evaluate whether this 
hypothesis correctly and accurately 
answers the question. First, provide 
a brief step-by-step analysis. Then, 
output a final confidence score between
0 and 100 indicating the probability 
that this hypothesis is the true answer.
Strictly format your final score within 
tags, exactly like this: <score>X</score>.
\end{PromptBox}

\subsection{Strategy details}
\label{appendix:strategies}

Direct is the rotation-0 observation of the cyclic permutation run, and Reasoning Direct that of the Reasoning Cyclic run; neither is a separate generation. Cyclic permutation and Reasoning Cyclic issue one request per cyclic rotation of the options ($n$ requests for an $n$-option question). Each parsed letter is mapped back to the canonical order, and the final answer is the majority vote, with ties resolved by the canonical tie-break (Appendix~\ref{appendix:ihs}). Two-stage and Reasoning Two-stage make two requests per question. The configured fallback to a direct prompt was never triggered in any canonical run. Hidden+Semantic makes no new request; it re-matches the saved Two-stage Stage 1 answer with the semantic-matching cascade. Visible+Semantic makes one request; Visible+LLM reuses that request's output and adds one LLM-matcher request. IHS makes $n$ requests per question.

\subsection{Semantic-matching pipeline}
\label{appendix:semantic}

Hidden+Semantic and Visible+Semantic map free text to an option with a three-stage cascade. Free text and option texts are first normalized (lower-cased, whitespace-collapsed). The stages are: (i) exact match; (ii) unique containment, where an option's text is contained in the free text or vice versa, applied only when the contained string has at least 4 characters and exactly one option qualifies; (iii) cosine similarity between \texttt{all-MiniLM-L6-v2} sentence embeddings of the free text and each option, taking the argmax among options with similarity $\geq 0.30$. If no option reaches the threshold, the question is unscored; no option is forced. Ties at any stage use the canonical tie-break. The Visible+Semantic results were produced by replaying the saved generations through the cascade without new provider requests.

\subsection{IHS scoring and deterministic tie-breaking}
\label{appendix:ihs-impl}

For each option, the score is the last \texttt{<score>X</score>} match in the response, where $X$ is an integer or decimal. A score that is missing, non-finite or outside $[0,100]$ is invalid and excluded from the argmax. A question with no valid score is unscored. Score ties are resolved by the canonical tie-break over each option's canonical ID (Appendix~\ref{appendix:ihs}).

\subsection{Artifact Documentation}
\label{appendix:artifacts}

\paragraph{Licensing and intended use.} 
MMLU \citep{hendrycks2021mmlu} is distributed under the MIT license, while ARC-Challenge \citep{clark2018arc} is listed under CC BY-SA 4.0. Qwen2.5-7B-Instruct \citep{qwen2025qwen25technicalreport} uses Apache 2.0, while Llama 3.1 8B Instruct \citep{grattafiori2024llama3herdmodels} is distributed under the Llama 3.1 Community License. The semantic matcher uses all-MiniLM-L6-v2 \citep{minilmModelCard}, distributed under Apache 2.0. GPT-4.1 mini and Claude Haiku 4.5 were accessed through their respective commercial APIs. All artifacts are used for research evaluation, rather than model training or deployment.

\paragraph{Data considerations.}
We use existing public benchmark datasets and do not collect new participant information. No additional manual or automated screening for personally identifying information or offensive content was conducted. We therefore rely on the original datasets' documentation for their data characteristics.

\paragraph{Computational resources.}
All experiments were inference-only. Qwen2.5-7B-Instruct and Llama-3.1-8B-Instruct have approximately 7B and 8B parameters, respectively. The parameter counts and provider-side hardware configurations of the proprietary models are not publicly disclosed. Most experiments used hosted APIs, with local verification performed using the configurations described in Appendix~\ref{appendix:results}.

\subsection{Decoding, seeds and deployment}
\label{appendix:deployment}

\paragraph{Decoding.} All conditions use temperature 0.0 and a maximum of 1024 new tokens; there is no method-specific token budget. The run seed is 42. It fixes question selection and the deterministic tie-break; no provider-side sampling seed is sent.

\paragraph{Benchmarks.} MMLU uses a frozen manifest of 1140 test questions (57 subjects $\times$ 20, seed 42). ARC-Challenge uses all 1172 test questions.

\paragraph{API deployments.} \texttt{gpt-4.1-mini-\allowbreak 2025-04-14} (OpenAI) and \texttt{claude-haiku-4-5-\allowbreak 20251001} (Anthropic) were called directly. Cyclic permutation, Reasoning Cyclic and IHS used the providers' batch APIs; the other methods were called synchronously with concurrency 10. \texttt{qwen/\allowbreak qwen-2.5-7b-instruct} was served via OpenRouter, pinned to the Phala upstream with fallbacks disabled; all methods ran synchronously. The serving precision of this deployment is not documented. \texttt{meta-llama/\allowbreak llama-3.1-8b-instruct} was served via OpenRouter, pinned to DeepInfra (FP8), for Two-stage, Visible+Semantic and Reasoning Two-stage. Cyclic permutation, Reasoning Cyclic and IHS used DeepInfra's batch endpoint with \texttt{meta-llama/\allowbreak Meta-Llama-3.1-8B-\allowbreak Instruct-Turbo}. Visible+LLM used the same pinned OpenRouter routes as Two-stage (DeepInfra for \texttt{meta-llama/\allowbreak llama-3.1-8b-instruct}, Phala for \texttt{qwen/\allowbreak qwen-2.5-7b-instruct}), called synchronously. 

\end{document}